\documentclass[journal, 10pt, twocolumn]{IEEEtran}

\usepackage[usenames, dvipsnames]{color}
\usepackage{citesort}
\usepackage{pifont}
\newcommand{\cmark}{\ding{51}}%
\ifCLASSINFOpdf
  \usepackage[pdftex]{graphicx}
	\usepackage{epstopdf}
\else
  \usepackage[dvips]{graphicx}
\fi
\usepackage[cmex10]{amsmath}
\usepackage{algorithmic}
\usepackage{amssymb}
\usepackage{array}
\usepackage{mdwmath}
\usepackage{mdwtab}
\usepackage{eqparbox}
\usepackage{multirow}
\usepackage{fixltx2e}
\usepackage{enumerate}
\usepackage{url}
\usepackage{tabularx}
\usepackage{verbatim}
\usepackage{color}

\newcommand{\chg}[1]{{\color{black}#1}}
\newcommand{\new}[1]{{\color{black}#1}}

\newcommand{\Cmb}{\mathbb{C}}

\newcommand{\diag}[1]{\text{diag}(#1)}
\newcommand{\mbf}[1]{\boldsymbol{#1}}
\newcommand{\mbb}[1]{\mathbb{#1}}

\newcommand{\xb}{\mbf{x}}

\newcommand{\Ab}{\mbf{A}}

\newcommand{\ub}{\mbf{u}}
\newcommand{\ubh}{\hat{\mbf{u}}}
\newcommand{\vb}{\mbf{v}}
\newcommand{\eb}{\mbf{e}}
\newcommand{\Db}{\mbf{D}}
\newcommand{\Dbh}{\hat{\mbf{D}}}
\newcommand{\Ub}{\mbf{U}}

\newcommand{\Ib}{\mbf{I}}
\newcommand{\sumi}{\sum_{i=1}^{N}}
\newcommand{\Sb}{\mbf{S}}
\newcommand{\Gb}{\mbf{G}}
\newcommand{\Bb}{\mbf{B}}
\newcommand{\bb}{\mbf{b}}
\newcommand{\zb}{\mbf{z}}
\newcommand{\sumj}{\sum_{j=1}^{M}}
\newcommand{\sumk}{\sum_{k=1}^{N}}
\newcommand{\suml}{\sum_{j=1}^{l}}

\newcommand{\Wb}{\mbf{W}}
\newcommand{\Wbh}{\hat{\mbf{W}}}
\newcommand{\Fbu}{\mbf{F}_{\boldsymbol{g}}}
\newcommand{\Fb}{\mbf{F}}

\newcommand{\Aset}{\left \{ \alpha_i \right \}}
\newcommand{\Dset}{\left \{ \Db_i \right \}}

\newcommand{\Rf}{\mathfrak{R}}

\begin{document}
%
\title{The Power of Complementary Regularizers: \\ Image Recovery \\ via Transform Learning and Low-Rank Modeling}
\author{Bihan~Wen, ~\IEEEmembership{Student Member,~IEEE,}~Yanjun~Li, ~\IEEEmembership{Student Member,~IEEE,} and~Yoram~Bresler,~\IEEEmembership{Fellow,~IEEE}
\thanks{This work was supported in part by the National Science Foundation (NSF) under grants CCF-1320953 and IIS 14-47879.}
\thanks{B. Wen, Y. Li, and Y. Bresler are with the Department of Electrical and Computer Engineering and the Coordinated Science Laboratory, University of Illinois, Urbana-Champaign, IL, 61801 USA e-mail: (bwen3, yli145, ybresler)@illinois.edu.}
}

\maketitle

\begin{abstract}
Recent works on adaptive sparse and on low-rank signal modeling have demonstrated their usefulness in various image/video processing applications. 
Patch-based methods exploit local patch sparsity, 
whereas other works apply low-rankness of grouped patches to exploit image non-local structures.
However, using either approach alone usually limits performance in image reconstruction or recovery applications.
In this work, we propose a simultaneous sparsity and low-rank model, dubbed STROLLR, to better represent natural images.
In order to fully utilize both the local and non-local image properties, we develop an image restoration framework using a transform learning scheme with joint low-rank regularization.
The approach owes some of its computational efficiency and good performance to the use of transform learning for adaptive sparse representation rather than the popular synthesis dictionary learning algorithms, 
which involve approximation of NP-hard sparse coding and expensive learning steps.
We demonstrate the proposed framework in various applications to image denoising, inpainting, and compressed sensing based magnetic resonance imaging. 
Results show promising performance compared to state-of-the-art competing methods.
\end{abstract}

\begin{IEEEkeywords}
Sparse representation, Image denoising, Image inpainting, Image Reconstruction, Block matching, Collaborative filtering, Machine learning. 
\end{IEEEkeywords}

\IEEEpeerreviewmaketitle

\section{Introduction} \label{sec1}

Image reconstruction refers to the process of forming an image from a collection of measurements.
Despite today's vast improvement in camera sensors, digital images are often still corrupted by severe noise in low-light conditions.
Furthermore, in modern computed imaging applications, in order to reduce the system complexity, data-acquisition time, or radiation dose, 
it is usually required to reconstruct high-quality images from incomplete or corrupted measurements.
Under such settings, image reconstruction corresponds to a challenging inverse problem.
We aim to estimate the underlying image $\mbf{x}$ from its degraded / noisy measurement $\mbf{y}$, which has the general form of $\mbf{y} = \Ab\, \mbf{x} + \eb$, where $\Ab$ and $\eb$ denote the sensing operator and additive noise, respectively.
This framework encompasses various important problems, including image denoising, deblurring, inpainting, super-resolution, compressed sensing (CS), and more advanced linear computed imaging modalities.
For such problems, and especially for those that are ill-posed, an effective regularizer is key to a successful image reconstruction algorithm.
Most of the popular methods take advantage of either sparsity or non-local image structures.

\begin{figure}[!t]
\begin{center}
\begin{tabular}{c}
\hspace{-0.1in}
\vspace{-0.1in}
\includegraphics[height=1.68in]{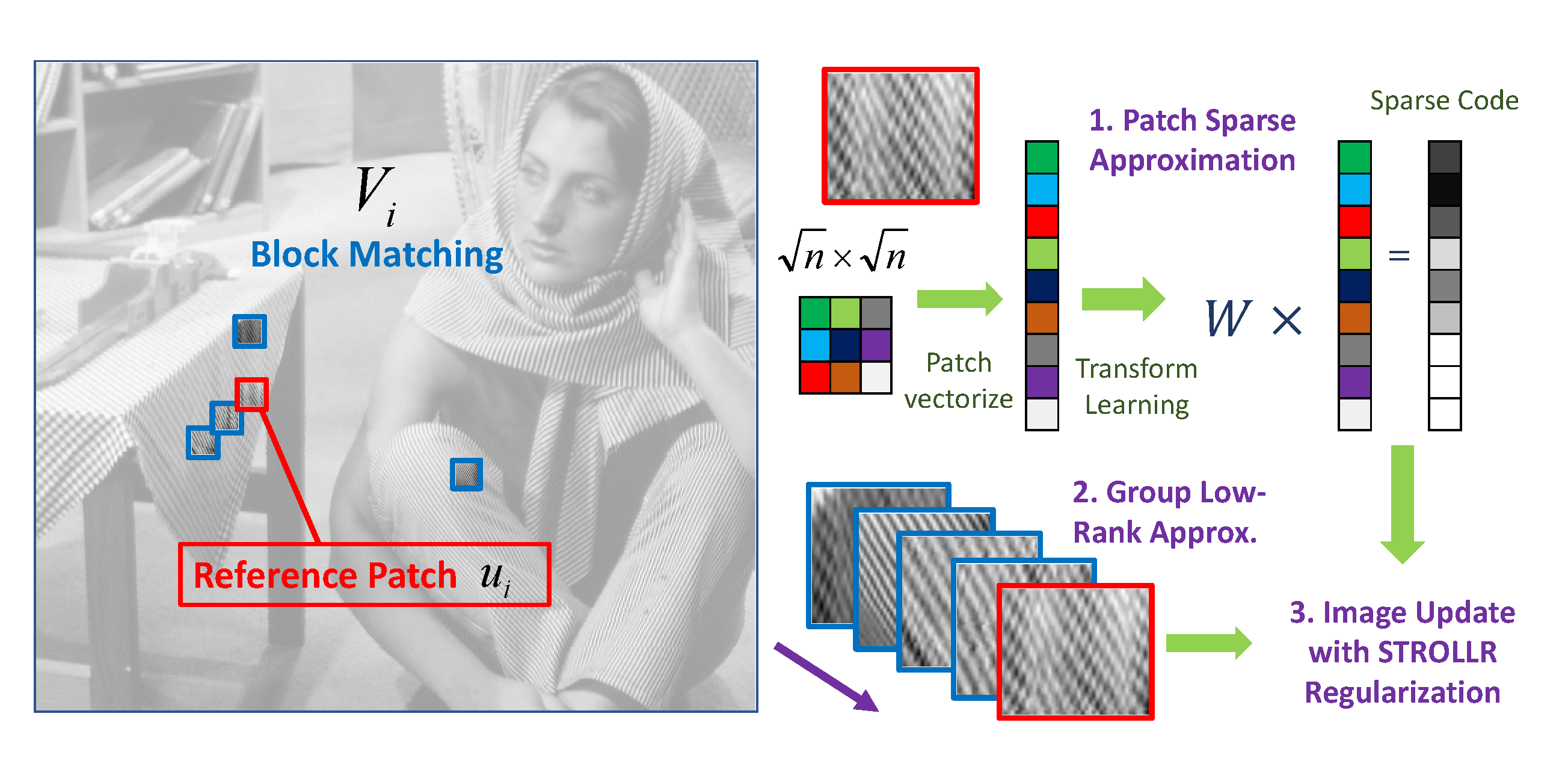}\\
\end{tabular}
\vspace{0.05in}
\caption{The STROLLR model for natural images, using group low-rankness and patch sparsity simultaneously.}
\label{frame}
\end{center}
\vspace{-0.2in}
\end{figure}

\subsection{Sparsifying Transform Learning} \label{sec1a}
It is well-known that natural images contain local structures, such that image patches are typically sparsifiable or compressible under certain transforms, or over certain dictionaries. 
Early works exploited the sparsity in a fixed transform domain, e.g. discrete cosine transform (DCT) \cite{Dabov2007,yu2011dct} and wavelets \cite{chang2000adaptive}.
Comparing to fixed sparse models, recent works have shown that the data-driven adaption of sparse signal models leads to promising results in various image recovery problems \cite{elad,elad2,Mairal2009,sabres3,octobos,wen2017sparsity}.
Among them, synthesis dictionary learning \cite{elad,elad2} is the most popular adaptive sparse modeling technique. 
However synthesis model based methods typically involve the approximate solution of an NP-hard sparse coding step \cite{npa}. 
The widely used approximate methods \cite{omp,elad2} are not efficient for large-scale problems.

As an alternative, the \textit{transform model} provides cheap and exact sparse coding. It models a signal $\ub \in \Cmb^{n}$ as approximately sparsifiable using a transform $\Wb \in \Cmb^{m \times n}$, i.e., $ \Wb\, \ub = \alpha + \eb $, where $\alpha \in \Cmb^{m} $ is sparse, and $\eb$ is a small transform-domain modeling error\footnote{This feature distinguishes the transform model from the related analysis dictionary model~\cite{rubinstein2013analysis}, in which the modeling error is in the signal domain.}.
A key advantage of this model over the synthesis (as well as the analysis) dictionary model, is that for a given transform $\Wb$, 
the optimal sparse code $\mbf{x}$ of sparsity level $s$ minimizing the modeling error $\| \mbf{e}\|_2$ is obtained \emph{exactly and cheaply} by simple thresholding of $\mbf{W} \mbf{u}$ to its $s$ largest magnitude components.
Recent works on sparsifying transform learning proposed efficient learning algorithms with convergence guarantees \cite{sabres3,octobos}, which turn out to be advantageous in applications including video denoising \cite{vidosat}, magnetic resonance imaging (MRI) \cite{MRIoctobos,frist,ravishankar2015efficient}, and computational tomography (CT) \cite{pfisterCT}, with state-of-the-art performances.

\subsection{Low-Rank Approximation} \label{sec1b}
Apart from sparsity, images also contain non-local structures such as self-similarity:
patches are typically similar to other non-local structures within the same image.
Recent image restoration algorithms investigated the grouping of similar patches, and exploited the correlation within each group \cite{Dabov2007,Buades2005,Mairal2009,Dong2011,Dong2013,Zhang2014,gu2017weighted,yoon2014motion,zha2017image}.
Among them, the algorithms based on low-rank (LR) modeling have demonstrated superior performance in image recovery tasks \cite{Dong2013,yoon2014motion,gu2017weighted}.
A successful approach of this nature comprises four major steps:
\begin{enumerate}
\item For each overlapping image patch $\mbf{u}_i$, apply block matching to find its similar patches.
\item Construct a data matrix $\Ub_i$ whose columns are the vectorized patches closest to $\mbf{u}_i$.
\item Denoise $\Ub_i$ by calculating its low-rank approximation. 
\item Aggregate the denoised $\Ub_i$'s to form the image estimate.
\end{enumerate} 
Step (3) aims to find a low-rank matrix $\Db_i$ to approximate each $\Ub_i$ by minimizing $\left \| \Ub_i - \Db_i \right \|_{F}^{2} \, + \, \Rf(\Db_i)$, where $\Rf(\Db_i)$ is the rank penalty. 
Similar to adopting sparsity-promoting ``norms'' as penalties in sparse coding, several types of norms have been introduced to impose low-rankness, including nuclear norm, Schattern p-norm \cite{yoon2014motion}, weighted nuclear norm \cite{gu2017weighted}, etc.


\subsection{Methodologies and Contributions} \label{sec1c}
In summary, transform learning, and low-rankness in groups of similar patches, respectively, capture the sparsity, and non-local self-similarity in natural images.
Each of them has been applied as an effective regularizer in various image restoration algorithms. 
However, to the best of our knowledge, no 2D image recovery algorithm has to date utilized both transform learning and low-rank approximation jointly.
In this work, we propose a flexible Sparsifying TRansfOrm Learning and Low-Rank (STROLLR) model. 
Instead of using nuclear norms, we directly minimize the rank of data matrix as the regularization term, which leads to a more efficient algorithm.
Our methodology and results are summarized as follows:

\begin{itemize}
	\item We propose a flexible STROLLR model that combines the adaptive transform sparsity of image patches and the low-rankness of data matrices formed by block matching (BM), thus taking full advantage of both the sparsity and non-local self-similarities in natural images.

 	\item We propose an image recovery framework using STROLLR learning with variational formulations. 
	This enables the solution of various inverse problems in the same framework, including image denoising, inpainting, CS MRI, etc.
	
	\item We develop efficient block coordinate descent algorithms solving a wide range of inverse problems using STROLLR learning. Each step of our proposed algorithms has a simple and exact solution. We evaluate our proposed algorithms over a set of testing datasets, demonstrating competitive performance compared to the leading published methods, including recent methods based on deep neural networks.
	
\end{itemize}

This paper is an extension of our previous conference work \cite{wen2017sparsity} that briefly investigated STROLLR algorithm.
While the image recovery scheme in \cite{wen2017sparsity} can only handle inverse problems with patch-wise sensing operator $\Ab$, such as image denoising and inpainting, here we generalize to applications with image-wise $\Ab$, such as CS MRI.
Furthermore, the algorithms in \cite{wen2017sparsity} learned sparsifying transforms over 2D patches. 
Similar to the extension in \cite{wen2017joint}, we train a 3D sparsifying transform over groups of highly correlated patches, which leads to improved image recovery performance.
We provide detailed experimental results using the proposed STROLLR algorithms, with extensive evaluation over several datasets, and comparison to related competing methods.

\vspace{-0.05in}
\subsection{Organization} \label{sec1d}

The rest of the paper is organized as follows. 
Section \ref{related} summarizes the related works on image recovery problems, including image denoising, inpainting, and CS MRI.
Section \ref{sec2} introduces the proposed STROLLR model and the learning formulation. 
Then, Section \ref{sec3} describes the image restoration problem based on STROLLR, as well as a simple algorithm using block coordinate descent. 
Section \ref{sec4} discusses three image restoration / recovery applications, namely image denoising, inpainting, and CS MRI.
Section \ref{sec5} demonstrates the behavior and promise of the
proposed algorithms for various image restoration / recovery applications over several standard datasets. 
Section \ref{sec6} concludes with proposals for future work.


\section{Related Works} \label{related}

\subsection{Image Denoising} \label{sec:denoising}

Image denoising is one of the most important problems in image processing and low-level computer vision.
It is dedicated to recovering high-quality images from their corrupted measurements, which also improves robustness in various high-level vision tasks \cite{liu2017image}.
The image denoising algorithms can be divided into \textit{internal} and \textit{external} methods \cite{burger2013learning,zontak2011internal,chen2015external,wang2015learning}.

\textit{Internal} methods make use of only the noisy image to be reconstructed.
Classical algorithms exploit image local structures using total variation (TV) \cite{rudin1992nonlinear,beck2009fast}, or sparsity in fixed transforms \cite{chang2000adaptive,Dabov2007,yu2011dct}.
Noise is reduced by various types of coefficient shrinkage, e.g. sparse coding of the compressed representation \cite{chang2000adaptive,beck2009fast}.
More recently, data-driven approaches demonstrated promising results in image sparse modeling, including dictionary learning and transform learning, and thus lead to better denoising performance compared to those using analytical transforms \cite{Mairal2009,elad2,octobos,wen2017sparsity}.
Beyond these local structures, images also contain non-local structures, such as non-local self-similarity.
Recent works proposed to group similar image patches, and denoise each group explicitly by applying collaborative filtering \cite{Dabov2007,Buades2005}, group-based sparsity \cite{Mairal2009,Zhang2014,dong2015image,zha2017image}, joint sparsity \cite{Dong2011}, low-rankness \cite{Dong2013,Dong2014,yoon2014motion,Li2015,gu2017weighted}, etc.
Yin et al~\cite{yin2017tale} proposed to use the row and column spaces of the stacked patch matrix to capture the local and non-local properties of the image, respectively, representing them by ``convolution framelets" that capture both properties simultaneously. 
This formulation was used to interpret and improve upon the low dimensional manifold model (LDMM)~\cite{osher2017low}. 
However, in this formulation, the local structure is represented in a linear way (not by sparsity).
This is different from our proposed approach, in which sparsity and low-rankness are simultaneously imposed in the image model.
Table \ref{Tab:compareDenoising} summarizes the key attributes of some of the aforementioned related image denoising algorithm representatives, as well as the proposed STROLLR method.

In addition to exploiting image internal structures, \textit{external} methods learn the image model using a corpus of clean training images.
The well-known fields of experts (FoE) method \cite{roth2009fields}, and the EPLL algorithm \cite{zoran2011learning} proposed to restore an image using a probabilistic model for image patches, which is learned on a corpus of clean image patches.
The PGPD \cite{xu2015patch} and PCLR \cite{chen2015external} algorithms construct Gaussian Mixture Models (GMM) using patch groups from a training corpus, with additional sparsity and low-rank regularizers, respectively, which achieved improved denoising results.
More recently, deep neural networks (DNN) have demonstrated remarkable potential to learn image models from training dataset with an end-to-end approach \cite{burger2012image,schmidt2014shrinkage,chen2017trainable,liu2017image}. 
The shrinkage field (SF) \cite{schmidt2014shrinkage} and trainable nonlinear reaction diffusion (TNRD) \cite{chen2017trainable} networks unrolled iterative denoising algorithms, that are based on analysis sparse models.
Besides networks derived by unrolling an iterative algorithm for a variational formulation, other popular neural networks structures, such as fully connected networks (FCN), convolutional neural networks (CNN), recurrent neural networks (RNN), and U-Net, have been applied to image restoration with state-of-the-art results \cite{burger2012image,zhang2017beyond,liu2017image,liu2018non}.

Although \textit{external} methods often demonstrate superior denoising performance, they are supervised algorithms that require training on a corpus of images with distribution similar to  the images to be denoised. 
It is expensive, or sometimes impossible to obtain a reliable training set of this kind in applications, such as remote sensing, biomedical imaging, scientific discovery, etc.  
In this work, we restrict our attention to \textit{internal} image denoising algorithms, and leave the combination with external methods to future work.

\begin{table}[t!]
\centering
\fontsize{9}{12pt}\selectfont
\begin{tabular}{|c|c|c|c|c|c|}
\hline
\multirow{2}{*}{\textbf{ Methods }} & \multicolumn{2}{c|}{Sparse Model} & Collab. & Joint  & Low-  \\

\cline{2-3}
 & Fixed & Learned & Filtering & Sparse & Rank  \\
\hline
ODCT & \cmark & & & & \\
\hline
KSVD &  & \cmark & & & \\
\hline
OCTOBOS &  & \cmark & & & \\
\hline
NLM &  &  & \cmark & & \\
\hline
BM3D & \cmark &  & \cmark & &  \\
\hline
GSR &  & \cmark & & \cmark & \\
\hline
SAIST &  & & & & \cmark \\
\hline
STROLLR &  & \cmark & & & \cmark\\
\hline
\end{tabular}
\caption{Comparison between internal image denoising methods, including ODCT \cite{elad2}, KSVD \cite{elad}, OCTOBOS \cite{octobos}, NLM \cite{Buades2005}, BM3D \cite{Dabov2007}, GSR \cite{Zhang2014}, SAIST \cite{Dong2013}, and STROLLR (this work). }
\label{Tab:compareDenoising}
\vspace{-0.2in}
\end{table}

\subsection{Image Inpainting} \label{sec:inpainting}

The term \textit{image inpainting} \cite{bertalmio2000image} refers to the process of recovering the missing pixels in an image.
The inpainting problem is encountered in many image applications, including image restoration, editting (e.g., object removal), texture synthesis, content-aware image resizing (e.g., image enlargement), etc.
In this paper, we restrict to inpainting problems in image recovery application, in which the missing region is generally small (e.g., random pixels missing), and the goal of inpainting is to estimate the underlying complete image.

Similar to denoising, successful image inpainting algorithms exploit sparsity or non-local image structures.
Popular inpainting methods exploit the structure in a local neighborhood of the missing pixels.
Bertalmio et al pioneered the work based on partial differential equations (PDEs) to propagate image local structures from known region to missing pixels \cite{bertalmio2000image}. 
Besides, classical inpainting algorithms also applied TV as image regularizers \cite{ballester2001filling,levin2003learning}.
Sparse priors have also been applied in inpainting problems, assuming the unknown and known part of the image share the same sparse model \cite{elad2010book}.
Recent works proposed image inpainting methods based on patch sparsity, using dictionary learning \cite{guleryuz2006nonlinear,xu2010image} and transform learning \cite{wen2017frist,wen2017sparsity}, demonstrating promising performance.
On the other hand, \textit{non-local} methods group similar image components and exploit their correlation.
Ram et al. \cite{ram2013image} proposed to order image patches in a shortest path followed by collaborative filtering for inpainting.
Li \cite{li2011image} proposed to iteratively cluster similar patches and reconstruct each cluster via sparse approximation.
Jin and Ye~\cite{jin2015annihilating} proposed inpainting algorithm using low-rank Hankel structured matrix completion.
More recently, non-local algorithms such as GSR \cite{Zhang2014} applied dictionary learning within each group of similar patches, for improved sparse representation in inpainting problems.
We refer the readers to a comprehensive review of various recent image inpainting approaches \cite{guillemot2014image}.

\subsection{Compressed Sensing MRI} \label{sec:MRI}

\begin{table}[t!]
\centering
\fontsize{9}{12pt}\selectfont
\begin{tabular}{|c|c|c|c|c|c|}
\hline
\multirow{2}{*}{\textbf{ Methods }} & \multicolumn{3}{c|}{Sparse Model} & Non- & Super-\\
\cline{2-4}
 & Fixed & Direct. & Learned  & Local  & vised \\
\hline
Sparse MRI & \cmark & & & & \\
\hline
PBDWS & \cmark & \cmark &  & & \\
\hline
DLMRI & & & \cmark & & \\
\hline
TLMRI & & & \cmark & & \\
\hline
FRIST- & &  \multirow{ 2}{*}{\cmark} & \multirow{2}{*}{\cmark} & & \\
MRI &  & & & & \\
\hline
PANO & \cmark &  & & \cmark & \\
\hline
ADMM-Net & & & \cmark &  & \cmark \\
\hline 
STROLLR- & & & \multirow{2}{*}{\cmark}  & \multirow{ 2}{*}{\cmark} & \\
MRI &  & & & & \\
\hline
\end{tabular}
\caption{Comparison between various MRI reconstruction methods, including SparseMRI \cite{lustig2007sparse}, PBDWS \cite{ning2013magnetic}, DLMRI \cite{ravishankar2011mr}, TLMRI \cite{ravishankar2015efficient}, FRIST-MRI \cite{wen2017frist}, PANO \cite{qu2014magnetic} and STROLLR-MRI (this work). }
\label{Tab:compareMRI}
\vspace{-0.2in}
\end{table}

In modern imaging applications, the image recovery problem from the sparsely sampled measurements is often ill-posed. 
A popular approach to recover high-quality images, is to use regularizers based on image priors that penalize the undesired solutions \cite{geethanath2013compressed}.
In this paper, we focus on one popular example of an ill-posed imaging problem, Compressed sensing (CS) MRI.
CS techniques enable accurate MRI reconstruction from undersampled k-space (i.e., Fourier domain) measurements, by utilizing image sparsity.
Popular CS MRI methods exploit either sparsity, or non-local self-similarity of the image. 
Here we survey several popular algorithms that are related to our proposed STROLLR-MRI.
Comprehensive reviews can be found in \cite{lustig2008compressed,jaspan2015compressed,geethanath2013compressed}.

To exploit image sparsity, Lustig et al \cite{lustig2007sparse} proposed the Sparse MRI method, which uses wavelets and total variation regularization.
Compared to such analytical transforms, adaptively learned transforms, or dictionaries have proved to be more effective for image modeling \cite{elad2010book,octobos,ravishankar2015efficient}.
Ravishankar and Bresler \cite{ravishankar2011mr,ravishankar2015efficient} utilized dictionary learning (DL) and transform learning (TL) for MR image reconstruction achieving superior results.
In other work, the PBDWS algorithm \cite{ning2013magnetic} used partially adaptive Wavelets to form an MR image regularizer that exploited the patch-based geometric directions.
More recently, the FRIST-MRI method \cite{wen2017frist} proposed to learn a sparsifying transform that is invariant to image patch orientations.
On the other hand, non-local methods exploit the image self-similarity for high-quality MRI reconstruction.
The PANO algorithm \cite{qu2014magnetic} used BM to group similar image patches, and applied the 3D Haar wavelet transform to model each group.
Furthermore, Yoon et al. \cite{yoon2014motion} proposed to approximate group-matched patches as low-rank.
More recently, Yang et al proposed ADMM-Net~\cite{sun2016deep} to unroll the well-known alternating direction method of multipliers (ADMM) algorithm \cite{boyd2011distributed} applied to a standard variational formulation with p-norm sparsity regularization, into a feed forward neural network.
ADMM-Net uses end-to-end training of linear operators that were fixed in the original variational formulation and ADMM algorithm. 
ADMM-Net achieved  state-of-the-art performance in CS MRI reconstruction
This approach requires supervised training, in which the training corpus and the sampling patterns need to have distributions similar to those of the latent MRI measurements to be reconstructed.
Table \ref{Tab:compareMRI} summarizes the major attributes of the aforementioned CS MRI algorithms, as well as our proposed method.




\section{The STROLLR Model and Image Recovery} \label{sec2}

We propose a general image recovery framework based on the STROLLR model for image regularization.
The goal is to recover an image  (in vectorized form) $\mbf{x} \in \Cmb^{p}$ from its degraded measurement $\mbf{y} \in \Cmb^{q}$ using the classical variational formulation
\begin{equation} 
\nonumber (\mathrm{P1})\;\;\; \hat{\mbf{x}} = \underset{\mbf{x}}{\operatorname{argmin}} \: \gamma^{F} \left \| \Ab \mbf{x} - \mbf{y} \right \|_{2}^{2} + \Rf_{strollr} (\mbf{x})\, ,
\label{framework:strollr}
\end{equation}
where $\gamma^{F} \left \| \Ab \mbf{x} - \mbf{y} \right \|_{2}^{2}$ is the image fidelity term with $\mbf{y}$ being the measurement under the sensing operator $\Ab \in \Cmb^{q \times p}$, \chg{and $\gamma^{F}$ being its weight.}
The structure of $\Ab$ varies in different image restoration problems. 
Here $\Rf_{strollr} (\mbf{x})$ is the STOLLR regularizer
which jointly imposes sparsity in a certain transform domain and group low-rankness of the data.
The proposed $\Rf_{strollr} (\mbf{x})$ is a weighted combination of non-local group low-rankness and sparsity penalties as follows
\begin{equation} 
\Rf_{strollr} (\mbf{x}) =  \, \gamma^{LR} \,\Rf_{LR} (\mbf{x}) \, + \, \gamma^S \, \Rf_S (\mbf{x}) \, ,
\label{eq:theReg}
\end{equation}
where $\gamma^{LR}$ and $\gamma^S$ are the corresponding weights.

The term $\Rf_{LR} (\mbf{x})$ of the STROLLR regularizer imposes a low-rank prior on groups of similar patches via a matrix rank penalty,
\begin{equation} 
\Rf_{LR} (\mbf{x}) = \underset{\{ \Db_{i} \}}{\operatorname{min}} \sumi \left \{ \left \| V_{i} \, \mbf{x} - \Db_{i}  \right \|_{F}^{2} + \theta^2 \, \mathrm{rank}(\Db_{i})  \right \} \, ,
\label{eq:lrReg}
\end{equation}
where $V_{i} : \mbf{x} \mapsto V_i \mbf{x} \in \Cmb^{n \times M} $ is a block matching (BM) operator. 
It takes $R_i \, \mbf{x}$ to be the reference patch, where $R_{i} \in \Cmb^{n \times p}$ extracts the $i$-th $n$-pixel overlapping patch of $\mbf{x}$.
\chg{The means of all overlapping patches are removed, and} $V_{i}$ selects $M$ patches $\{ \ub_j \}^i$ that are closest to $R_i \mbf{x}$ in Euclidean distance $\left \|  \ub_j - R_i \, \mbf{x} \right \|_{2}$.
There are $N$ patches in total extracted from the image $\mbf{x}$. 
The selected patches $\{ \ub_j \}^i$ are inserted into the columns of matrix $V_i \mbf{x}$ in ascending order of their Euclidean distance to $R_i \, \mbf{x}$. 
\chg{The removed means are added back once the patches are denoised via low-rank approximation.}
Computing the Euclidean distance between each patch pair, and sorting them can be very expensive for large images.
In practice, we set a square $\sqrt{Q} \times \sqrt{Q}$ pixel search window, which is centered at the reference patch. 
Only the overlapping patches within the search window are evaluated by the BM operator, assuming the neighborhood patches usually have higher spatial similarities.
The optimal $\Dbh_i$ is called the low-rank approximation of the matched block $ V_{i} \, \mbf{x} $.
The low-rank prior has been widely used to model spatially similar patch groups \cite{Dong2013,Ji2010,Dong2014,yoon2014motion}. 
Applying rank penalty leads to a simple low-rank approximation algorithm, which can be computed using singular value decomposition (SVD) and hard thresholding (see Section~\ref{sec3} for details).

The sparsity regularizer $\Rf_S (\mbf{x})$ assumes that a vectorized signal $\ub_i \in \Cmb^n$ is approximately sparsifiable by some transform $\Wb$ that is adapted to the data $\mbf{x}$. 
One way to construct the sparsifiable signals, is by using the vectorized 2D image patches \cite{wen2017sparsity}, i.e., $\ub_i \triangleq R_{i} \mbf{x}$.
Therefore, for a given transform $\mbf{W} \in \mbb{C}^{m \times n}$, the sparsity regularizer on 2D image patches is formulated as
\begin{align} \label{eq:sparReg}
\Rf_S^{2D} (\mbf{x}, \mbf{W}) = \underset{\{ \alpha_i \}}{\operatorname{min}} \sumi \left \{ \left \| \Wb R_i \mbf{x} - \alpha_i \right \|_{2}^{2} + \lambda^2 \left \| \alpha_i \right \|_{0} \right \}
\end{align}
where the $\ell_0$ ``norm" counts the number of nonzeros in each sparse vector $\alpha_i$.
Given the transform $\Wb$, the optimal $\hat{\alpha}_i$ is called the sparse code of $\ub_{i}$, which can be calculated easily by hard thresholding (see Section \ref{sec3}).

We further extend the sparsity regularizer to impose sparsity over 3D patches.
Instead of using $\ub_i \triangleq R_{i} \mbf{x}$, we construct the signals as $\ub_i \triangleq C_i \mbf{x} \in \Cmb^{nl}$.
The operator $C_i$ first maps the BM matrix $V_i \mbf{x}$  (with first column $R_i \mbf{x}$) to the sub-matrix formed by its first $l$ columns, and then vectorizes the sub-matrix (in column lexicographical order).
Therefore, for a given transform $\mbf{W} \in \mbb{C}^{m \times nl}$, the new sparsity regularizer $\Rf_S (\mbf{x}, \mbf{W})$ is formulated as
\begin{align} \label{eq:sparReg3D}
\Rf_S (\mbf{x}, \mbf{W}) = \underset{\{ \alpha_i \}}{\operatorname{min}} \sumi \left \{ \left \| \Wb \, C_i \mbf{x} - \alpha_i \right \|_{2}^{2} + \lambda^2 \left \| \alpha_i \right \|_{0} \right \} \, ,
\end{align}
where each sparse code $\alpha_i \in \Cmb^{m}$.
Instead of using analytical transforms, an adaptively learned $\Wb$ \cite{octobos,sabres3} provides superior sparsity, which serves as a better regularizer \cite{ravishankar2015efficient,vidosat,pfisterCT,frist,wen2017joint}. 
In the sparsity regularizer, the sparsifying transform is trainable, which is obtained by transform learning.
Generally, the sparsifying transform $\Wb$ can be overcomplete \cite{octobos} or square \cite{sabres3}, with different types of regularizers or constraints \cite{sabres3}. 
In this work, we restrict ourselves to learning a square (i.e., $m = nl$) and unitary transform (i.e., $\Wb^{H}\,\Wb = \Ib_{nl}$, where $\Ib_{nl} \in \Cmb^{nl \times nl}$ is the identity matrix)  \cite{sabres3}.
The sparsity regularization term in (\ref{eq:theReg}) is thus obtained as
\begin{align} \label{eq:learnedReg}
\Rf_S (\mbf{x}) = \underset{ \mbf{W} \in \mbb{C}^{nl \times nl} }{\operatorname{min}} \Rf_S (\mbf{x}, \mbf{W}) \;\; s.t.\; \mbf{W}^H \mbf{W} = \mbf{I}_{nl}.
\end{align} 
This optimization problem has a closed form solution requiring only the computation of the SVD of an ${nl \times nl}$ matrix, leading to highly efficient learning and image restoration algorithms \cite{frist,MRIoctobos,wen2017joint}. 

In order to recover the underlying image $\mbf{x}$, we use the STROLLR regularizer for image recovery. 
We combine ($\mathrm{P1}$) with (\ref{eq:theReg}), (\ref{eq:lrReg}), (\ref{eq:sparReg3D}) and (\ref{eq:learnedReg}), and pull the minimizations to the front.
Therefore, the STROLLR learning based image recovery problem is formulated as follows, 
\begin{align} 
\nonumber (\mathrm{P2})  \:  & \min_{\left \{\mbf{x}, \Wb, \left \{ \alpha_i, \Db_i \right \}\} \right \} }  \: \, \gamma^{F} \left \| \Ab \mbf{x} - \mbf{y} \right \|_{2}^{2} \\
\nonumber  & \;\;\;\;\;\;\;\; + \gamma^S \sumi \left \{ \left \| \Wb \, C_i \mbf{x} - \alpha_i \right \|_{2}^{2} + \lambda^2 \left \| \alpha_i \right \|_{0} \right \} \\
\nonumber &  \;\;\;\;\;\;\;\; +  \gamma^{LR} \sumi \left \{ \left \| V_{i} \, \mbf{x} - \Db_{i}  \right \|_{F}^{2} + \theta^2 \, \mathrm{rank}(\Db_i)  \right \} \\
\nonumber & \;\;\;\;\;\;\;\;\;\;\;\;\;\;\;\;\;\;\;\; s.t.\; \:  \Wb^{H}\Wb= \Ib_{nl}\; .
\end{align}



\section{Algorithm} \label{sec3}

\begin{figure} \label{algo:recon}
\begin{tabular}{p{8.3cm}}
\hline
\vspace{0.02in}
Algorithm \textbf{A1}: STROLLR-based Image Reconstruction
\vspace{0.02in}
\\
\hline
\vspace{0.02in}
\textbf{Input:} The measurement $\mbf{y}$. \\
\textbf{Initialize:} $\Wbh_{0} = \Wb_{0}$ (e.g., 2D DCT), and the image $\hat{\mbf{x}}_0$:
\textbf{For $\;t = 1, 2,..., T$ Repeat}
\begin{enumerate}
\item \textbf{Low-rank Approximation for all $i = 1, ... N$:} 
\begin{enumerate}
\item Form $\{ V_i \, \hat{\mbf{x}}_{t-1} \}$ using BM.
\item Compute the full SVD $\Gamma \, \text{diag}(\omega) \, \Upsilon ^{H} \leftarrow V_{i} \, \hat{\mbf{x}}_{t-1}$.
\item Update $\Dbh_{i} = \Gamma \, \text{diag}(H_{\theta}(\omega)) \, \Upsilon ^{H}$.
\end{enumerate}
\item \textbf{Sparse Coding:} 
$\hat{\alpha}_i =  H_{\lambda}(\Wbh_{t-1} \, R_i \, \hat{\mbf{x}}_{t-1})$.
\item \textbf{Transform Update:} 
Compute the full $\mathrm{SVD}$ $S\, \Sigma\, G^{H} \leftarrow \mathrm{SVD}(\sum_{i=0}^N\, (R_i\, \hat{\mbf{x}}_{t-1}) \hat{\alpha}_i)$, 
then update $\Wbh_t\, =\, G\, S^{H}$.
\item \textbf{Image Reconstruction:} 
Update $\hat{\mbf{x}}_t$ by solving the problem (\ref{eq:imageRecon}), with the specific $\Ab$.
\end{enumerate}
\textbf{End} \\
\textbf{Output:} The reconstructed image $\hat{\mbf{x}}_T$. \\
\hline
\end{tabular} 
\caption{The STROLLR image recovery algorithm framework.} 
\vspace{-0.05in}
\end{figure}

We propose a simple block coordinate descent algorithm framework to solve $(\mathrm{P2})$. 
Each iteration involves four steps: ($i$) low-rank approximation, ($ii$) sparse coding, ($iii$) transform update, and ($iv$) image reconstruction. 
For all applications under the general STROLLR image reconstruction framework (\ref{framework:strollr}), they follow the same STROLLR learning steps ($i$) - ($iii$).
The image initialization, and the image reconstruction step ($iv$) may vary in specific applications with different sensing operator $\Ab$'s. 

\subsubsection{Low-rank Approximation} 
For fixed $\mbf{x}$, Problem (P2) separates into subproblems that we solve for each low-rank approximant $\Db_i$ as,
\begin{align}
\hat{\Db}_i = \underset{\Db_i}{\operatorname{argmin}}\: \left \| V_{i} \, \mbf{x} - \Db_i  \right \|_{F}^{2} + \theta^2 \, \mathrm{rank}(\Db_i) \, .
\end{align} 
We form matrix $V_i \, \mbf{x} \in \Cmb^{n \times M}$ using BM within the $\sqrt{Q} \times \sqrt{Q}$ search window, which is centered at the $i$-th patch $\ub_i$. 
\chg{Note that the locations (i.e., indices) of the patches used to form $V_i \mbf{x}$ and $C_i \mbf{x}$ in each iteration of the algorithm are updated and stored, to be used for the image reconstruction step.
}
Let $\Gamma \, \text{diag}(\omega) \, \Upsilon ^{H} = V_{i} \, \mbf{x}$ be the full SVD, where the diagonal vector $\omega$ contains the singular values. Then the low-rank approximation $\hat{\Db}_i = \Gamma \, \text{diag}(H_{\theta}(\omega)) \, \Upsilon ^{H}$ is the exact solution.
Here the hard thresholding operator $H_{v}(\cdot)$ is defined as
\begin{equation} \label{thr}
 \nonumber \left ( H_{v} (\beta) \right )_r = \left\{\begin{matrix}
 0&, \;\;\left | \beta_r \right | < v \\
\beta_{j}  & ,\;\;\left | \beta_r \right | \geq v
\end{matrix}\right .
\end{equation}
where $\beta \in \Cmb^{n}$ is the input vector, $v$ is the threshold value, and the subscript $r$ indexes the vector entries. 

\subsubsection{Sparse Coding}
Given the initialization, or the update of image $\mbf{x}$ and transform $\Wb$, we solve Problem $(\mathrm{P2})$ for the sparse codes,
\begin{align}
\hat{\alpha_i} = \underset{\alpha_i}{\operatorname{argmin}}\: \left \| \Wb \, C_i \, \mbf{x} - \alpha_i \right \|_{2}^{2} + \lambda^2 \left \| \alpha_i \right \|_{0}\;\;\;\; \forall i\, , 
\end{align} 
which is the standard transform-model sparse coding problem. 
The optimal $\hat{\alpha}_i$ can be obtained using cheap hard thresholding, $\hat{\alpha}_i = H_{ \lambda } ( \Wb \, C_i\, \mbf{x} )$. 

\subsubsection{Transform Update} 
For fixed $x$ and $\{ \alpha_i \}$, we solve for unitary $\Wb$ in $(\mathrm{P2})$, which is equivalent to the following,
\begin{align}
\hat{\Wb} = \underset{\Wb}{\operatorname{argmin}} \sumi \left \| \Wb \, C_i \mbf{x} - \alpha_i \right \|_{2}^{2} \;\; s.t.\; \Wb^{H}\,\Wb=\Ib_n
\end{align} 
With the unitary constraint, the optimal $\hat{\Wb}$ has a simple and exact solution~\cite{sabres3}: denoting the full singular value decomposition (SVD) of $ \mbf{K} \triangleq \sumi (C_i \, \mbf{x}) \, \alpha_i^{H}$ as $\Sb \, \Sigma \, \Gb^{H}$, the transform update is $\hat{\Wb} = \Gb \, \Sb^{H}$.

\subsubsection{Image Reconstruction}
With updated $\Wb$, $\Dset$, and $\Aset$, we reconstruct the underlying image $\mbf{x}$ by solving the following problem,
\begin{align} \label{eq:imageRecon}
\nonumber \hat{\mbf{x}} & \:  = \:  \underset{\mbf{x}}{\operatorname{argmin}}\: \gamma^{F} \left \| \Ab \, \mbf{x} - \mbf{y} \right \|_{2}^{2} \\
& + \sumi \left \{  \gamma^S \left \| C_i \mbf{x} - \ubh_i \right \|_{2}^{2} + \gamma^{LR} \sumi \left \| V_i \, \mbf{x} - \Db_i \right \|_{F}^{2}\right \} .
\end{align}
Here $\ubh_i \triangleq \Wb ^{H} \hat{\alpha}_i$ denotes the reconstructed patches via the transform-model sparse approximation. 
Since the unitary $\Wb$ preserves the norm, we have $ \left \| C_i \mbf{x} - \ubh_i \right \|_{2}^{2} = \left \| C_i \mbf{x} - \Wb^{H} \alpha_i \right \|_{2}^{2} = \left \| \Wb C_i \mbf{x} - \alpha_i \right \|_{2}^{2}$.

The image reconstruction problem (\ref{eq:imageRecon}) is a least squares problem with solution given by the solution to the normal equation
\begin{align} \label{eq:normal}
\Bb \, \hat{\mbf{x}} \, = \,\zb \, ,
\end{align}
where the left and right sides of (\ref{eq:normal}) are defined as
\begin{align} \label{eq:B}
\Bb & \triangleq \Ab^{H} \Ab +  \gamma^S \sumi C_i^* C_i + \gamma^{LR} \sumi V_i^{*} V_i \\
\zb & \triangleq \mbf{y} + \gamma^S \sumi C_i^* \ubh_i + \gamma^{LR} \sumi V_i^{*} \Db_i.
\label{eq:Z}
\end{align}
Here $V_i^{*} : \mbb{C}^{n \times M} \rightarrow \mbb{C}^p$ and $C_i^* : \mbb{C}^{nl} \rightarrow \mbb{C}^p $ denote the adjoint operators of $V_i$ and $C_i$, respectively, which correspond to patch deposit operators.
In particular, $V_i^{*}$ takes an $n \times M$ matrix of $M$ patches, and ``deposits" the patches in their respective locations in a (vectorized) image.
Overlapping patches are added up where they overlap. 
A similar operation is performed by $C_i^*$ on a length-$nl$ vector, extracting $l$ length-$n$ consecutive subvectors  and depositing them as patches in their respective locations in a (vectorized) image.
In (\ref{eq:B}), both $\sumi C_i^* C_i$ and $\sumi V_i^{*} V_i$ are $p \times p$ diagonal matrices with $(j, j)$ elements equal to the total number of the patches in all the $C_i \mbf{x}$'s and $V_i \mbf{x}$'s that contain the $j$-th pixel, respectively.
In (\ref{eq:Z}), the image-size vector $\zb \in \Cmb^{p}$ is a weighted combination of noisy measurements, and the images formed by the sparse and low-rank approximations of patches.

There are different sensing operators $\Ab$ associated with various inverse problems, leading to different forms of $\Bb \in \Cmb^{p \times p}$, and to variations in the solutions to step ($iv$). 
Direct inversion of $\Bb$ is typically expensive, but there exist efficient inverses of $\Bb$ for some inverse problems.
Several exemplary applications will be discussed in Section \ref{sec5}.
The general image recovery algorithm using STROLLR learning is summarized as Algorithm \textbf{A1}.


\subsubsection{Computational Cost} 
In Algorithm \textbf{A1}, the computational cost for the STROLLR-based image recovery, excluding the image reconstruction step, is $O(NnQ + \operatorname{min}(NMn^2, NnM^2) + Nn^2l^2 + (Nn^2l^2 + n^3l^3))$ per iteration, corresponding to the steps of BM, low-rank approximation, sparse coding, and transform update, respectively,
 where the $Nn^2l^2$ term in the transform update step is the cost of forming matrix $\mbf{K}$, and the $n^3l^3$ corresponds to the cost of its SVD.
Here the number of patches $N$ scales similar to the image size $p \gg n$. 
Typically, the search window size needs to be sufficiently large, i.e., $Q \gg M, n$. The 3D patches are only formed by a small number of $l$ highly correlated 2D patches, i.e., $l^2 < M$.
Furthermore, the BM matrix size $M \gg 1$ and scales similar to $n$.
Thus, the cost of the STROLLR-based image recovery is dominated by the cost of BM and low-rank approximation steps, and scales as $O(NnQ + \operatorname{min}(NMn^2, NnM^2))$, i.e., as $O\left( n (Q + M \min (n, M ) ) \right)$ per image pixel per iteration.

Note that this cost analysis is based on full SVD, and naive BM algorithm.
Further cost reductions are possible by randomized SVD to obtain an approximate truncated SVD \cite{zhou2011godec,liberty2007randomized}, 
and by using fast data structures and algorithms for k-NN (with k=M nearest neighbors) to perform the block matching \cite{chen2001fast}.


\section{Image Recovery Applications} \label{sec4}

The STROLLR model is particularly appealing in recovery of natural images, as well as biomedical images.
In this section, we consider three such applications, namely image denoising, inpainting, and CS-based MRI.
Each corresponds to a specific sensing operator $\Ab$ in ($\mathrm{P2}$), thus leading to a different image reconstruction step in (\ref{eq:imageRecon}).


\subsection{Image Denoising} \label{sec41}

When $\Ab = \Ib_p$, we are solving the image denoising problem, which is one of the most fundamental inverse problems in image processing. 
The goal is to recover the image $\mbf{x}$ from its noisy measurement $\mbf{y} = \mbf{x} + \eb$, which is corrupted by noise vector $\eb$.
In the image denoising algorithm based on STROLLR learning, we initialize $\hat{\mbf{x}}_0 = \mbf{y}$.
The matrix $\Bb$ in (\ref{eq:B}) thus becomes
\begin{align} \label{eq:Bdenoising}
\Bb = &\, \text{diag}(\bb)  \triangleq \Ib_p +  \gamma^S \sumi C_i^* C_i + \gamma^{LR} \sumi V_i^{*} V_i \, ,
\end{align}
where $\Bb \in \Cmb^{p \times p}$ is a diagonal matrix with positive diagonal elements $\bb^j > 0 \,\, \forall j$.
Thus, with $z$ given by (\ref{eq:Z}), the denoised image has the closed form:
\begin{align} \label{eq:ImageUpdateDenoising}
\hat{\mbf{x}} = \Bb^{-1} \, \zb \, .
\end{align}
The inversion of the diagonal matrix $\Bb$ is simple, and the solution reduces to the pixel-wise division $\hat{\mbf{x}}_j = \zb_j / \bb^j\; \forall j = 1, ... , p$, which can be thought of as normalization to remove redundancy in $\zb$. 
The computational cost of this step is $O(p)$ per iteration, which is negligible compared to the cost of the other steps $(i) - (iii)$.
Thus, the computational cost of the STROLLR based image denoising algorithm is on par with various state-of-the-art method, such as BM3D \cite{Dabov2007}, SAIST \cite{Dong2013}, etc.

Though the proposed algorithm is designed to recover gray-scale images, it has a simple extension to color image denoising that exploits the correlation across the color channels.
There are algorithm modifications in both transform learning and low-rank approximation.
The red (R), green (G), and blue (B) channels
~\footnote{\chg{
The proposed denoising algorithm can be also applied to data in a different color space possibly weighting more a highly informative channel such as luminance.
}
} 
of each color patch are vectorized to form one training sample in transform learning.
In the low-rank approximation step, the BM operator $V_i$ selects the $M$ patches that have the minimum Euclidean distance to $R_i \mbf{x}$, summed over the three color channels. 
The block matched matrix $V_i \mbf{x}$ is formed by matched patches, in which the R, G, and B channels of each selected patches are in the adjacent columns.


\subsection{Image Inpainting} \label{sec42}
The goal of image inpainting is to estimate the missing pixel in an image. 
When $\Ab = \Phi \in \Cmb^{p \times p}$, the given image measurement is denoted as $\mbf{y} = \Phi \mbf{x} + \eb$, where the $\Phi$ is a diagonal binary matrix with zeros at the locations corresponding to missing pixels in $\mbf{y}$.
The vector $\eb$ denotes the additive noise on the available pixels. 
Similar to image denoising, we initialize $\hat{\mbf{x}}_0 = \mbf{y}$ in the STROLLR-based inpainting algorithm.

Similar to denoising, the inpainted image has a simple closed-form solution $\mbf{x} = \Bb^{-1} \, \zb$, where $\zb$ is again given by (\ref{eq:Z}) and  the normalization matrix is the diagonal matrix with positive diagonal elements:
\begin{align}
\Bb \triangleq \Phi +  \gamma^S \sumi C_i^* C_i + \gamma^{LR} \sumi V_i^{*} V_i \, .
\end{align}
The matrix inversion is again cheap pixel-wise division.

In the ideal case when the noise $\eb$ is absent, i.e., $\sigma = 0$, we replace the fidelity term $\left \| \Phi \, \mbf{x} - \mbf{y} \right \|_{2}^{2}$ with the hard constraint $\Phi \, \mbf{x} \; =  \; \mbf{y}$. 
The image reconstruction step becomes a constrained optimization problem,
\begin{align} \label{eq:idealInpainting}
\nonumber  \underset{\mbf{x}}{\operatorname{min}}\: \sumi & \begin{Bmatrix} \gamma^S \left \| C_i \mbf{x} - \ubh_i \right \|_{2}^{2}  + \gamma^{LR} \sumi \left \| V_i \, \mbf{x} - \Db_i \right \|_{F}^{2} \end{Bmatrix} \\
&\;\;\;\;\;\;\;\;\;\;\;\;\;\;\;\;\;\;\;\;\; s.t.\;\; \Phi \, \mbf{x} \; =  \; \mbf{y} \;.
\end{align}
Let $\Omega$ denote the subset of image pixels that are sampled by $\mbf{y}$, i.e., $\text{diag}(\Ab)_\Omega \neq 0$.
With the hard constraint on the pixels in $\Omega$, the closed-form solution $\hat{\mbf{x}}$ to (\ref{eq:idealInpainting}) becomes
\begin{equation}
 \hat{\mbf{x}}_{j} = \left \{ \begin{matrix}
 \zb_j / \bb_j & , \;\; j \notin \Omega \\
 \mbf{y}_j  & ,\;\; j \in \Omega
\end{matrix}\right .
\end{equation}
where the vectors $\zb$ and $\bb$ are defined as
\begin{align}
\text{diag}(\bb) & \triangleq \gamma^S \sumi C_i^* C_i + \gamma^{LR} \sumi V_i^{*} V_i \\
\zb & \triangleq \gamma^S \sumi C_i^* \ubh_i + \gamma^{LR} \sumi V_i^{*} \Db_i.
\end{align}
Similar to image denoising, the computational cost of the image reconstruction step in inpainting is also $O(p)$, i.e., $O(1)$ per pixel, per iteration, which is negligible relative to the costs of the other steps.
The computational cost of both the noisy and ideal STROLLR based image inpainting algorithms are on par with popular competing methods, such as GSR \cite{Zhang2014}.

\subsection{Magnetic Resonance Imaging} \label{sec43}

We propose an MR image reconstruction scheme based on STROLLR learning, dubbed STROLLR-MRI.
The sensing operator $\Ab = \mbf{F}_{\boldsymbol{g}} \in \Cmb^{q \times p}$ in (\ref{eq:imageRecon}) is the undersampled Fourier encoding matrix, composed of the $q$ rows of the unitary $p \times p$ 2D DFT matrix $\mbf{F}$ corresponding to the sampled locations in $k$ space. 
The selected rows are indicated by the positions of ones in the binary vector $\boldsymbol{g} \in \{0, 1 \}^p$.
The k-space measurement $\mbf{y} \in \Cmb^{q}$ has lower dimension, i.e., $q \ll p$, 
thus the MRI reconstruction is an ill-posed problem that requires an effective image regularizer. 
We can directly formulate STROLLR-MRI based on (P2), but this introduces several complications.
First, as the number of pixel references generated by BM is not homogeneous across the image, the $\mbf{B}$ matrix in (\ref{eq:B}) cannot be diagonalized by the DFT~\cite{ravishankar2011mr,ravishankar2015efficient,wen2017frist}, 
making direct inversion of $\mbf{B}$ impractical for MRI, requiring to perform the image update step (9) by iterative optimization methods, such as conjugate gradients at considerably higher computation cost. 
Second, as image content and therefore the matches by BM are
updated from one iteration to the next, the structure of the cost function will usually vary and affects the algorithm convergence~\cite{yoon2014motion}.


Instead, we propose to normalize the weights of patches that appear in multiple BM groups $V_i \, \mbf{x}$ or $C_i \, \mbf{x}$ by the number of their appearances, so that all patches exert similar influence on the regularizer.
We initialize the image by the so-called zero-filled DFT inverse, $\hat{\mbf{x}}_0 = \mbf{F}_{\boldsymbol{g}}^H \mbf{y}$ in the STROLLR-based MRI algorithm.
The STROLLR-MRI image reconstruction step (\ref{eq:imageRecon}) is replaced by,
\begin{align} \label{eq:mrirecon}
\nonumber \hat{\mbf{x}} \: = \: & \underset{\mbf{x}}{\operatorname{argmin}}\: \left \| \Fbu \, \mbf{x} - \mbf{y} \right \|_{2}^{2} \\
\nonumber + & \gamma^S \sumi \suml \frac{1}{P_{i, j}} \left \| (C_i \, \mbf{x})_j - \ubh_{i, j} \right \|_{2}^{2}  \\
+ & \gamma^{LR} \sumi \sumj \frac{1}{L_{i, j}} \left \| \, (\, V_i \, \mbf{x} \, )_j - \Db_{i, j}\,  \right \|_{2}^{2} \;\; .
\end{align}
Here $(\, V_i \, \mbf{x} \, )_j$ and $\Db_{i, j}$ denote the $j$-th column of the matrices $V_i \, \mbf{x}$ and $\Db_i$, respectively.
The weight $L_{i, j}$ equals the number of times that the $j$-th column of $V_i \, \mbf{x}$ appears in all $\begin{Bmatrix} V_i \, \mbf{x} \end{Bmatrix}_{i=1}^N$.
Similarly, we use $(C_i \mbf{x})_j \in \Cmb^n$ and $\ubh_{i, j} \in \Cmb^n$ to denote the $j$-th block of $C_i \mbf{x}$ and $\ubh_i \in \Cmb^{n}$, respectively.
The weight $P_{i, j}$ equals the number of times that the $j$-th block of $C_i \, \mbf{x}$ appears in all $\begin{Bmatrix} C_i \, \mbf{x} \end{Bmatrix}$.
Define the sets $\Delta_k = \begin{Bmatrix} (i, j)\; | \; (\,C_i \, \mbf{x} \,)_j = R_k \, x \end{Bmatrix}$, and $\Gamma_k = \begin{Bmatrix} (i, j)\; | \; (\,V_i \, \mbf{x} \,)_j = R_k \, \mbf{x} \end{Bmatrix}$, which indicate the indices $(i, j)$ where patch $R_k \, \mbf{x}$ appears in $C_i \, \mbf{x}$ and $V_i \, \mbf{x}$, respectively.
As all wrap-around patches are used as reference patches, each $R_k \mbf{x}$ appears at least once in $\begin{Bmatrix} V_i \mbf{x} \end{Bmatrix}$ and $\begin{Bmatrix} C_i \mbf{x} \end{Bmatrix}$, i.e., $\begin{vmatrix} \Delta_k \end{vmatrix} \geq 1$ and $\begin{vmatrix} \Gamma_k \end{vmatrix} \geq 1$, respectively.
Thus, each $R_k \, \mbf{x}$ can be represented as
\begin{align} \label{eq:intermediate2}
& R_k \, \mbf{x} \: = \frac{1}{\begin{vmatrix} \Delta_k \end{vmatrix}} \sum_{(i, j) \in \Delta_k} (C_i \, \mbf{x})_j =  \frac{1}{\begin{vmatrix}  \Gamma_k \end{vmatrix}} \sum_{(i, j) \in \Gamma_k} (V_i \, \mbf{x})_j \, .
\end{align}
Using (\ref{eq:intermediate2}) and the fact that $P_{i,j} = \begin{vmatrix} \Delta_k \end{vmatrix}$ for $(i,j) \in \Delta_k$, and $L_{i,j} = \begin{vmatrix} \Gamma_k \end{vmatrix}$ for $(i,j) \in \Gamma_k$, 
defining $\tilde{\mbf{u}}_k \triangleq \frac{1}{\begin{vmatrix} \Delta_k \end{vmatrix}} \sum_{(i, j) \in \Delta_k} \hat{\mbf{u}}_{i, j}$ and $\mbf{d}_k \triangleq \frac{1}{\begin{vmatrix} \Gamma_k \end{vmatrix}} \sum_{(i, j) \in \Gamma_k} \Db_{i, j}$, and dropping terms independent of $\xb$, (20) simplifies to 

\begin{align} \label{eq:MRIsimpleForm}
\nonumber \hat{\mbf{x}} \: = \: & \underset{\mbf{x}}{\operatorname{argmin}}\: \left \| \Fbu \, \mbf{x} - \mbf{y} \right \|_{2}^{2} \;\; + \\
& \sumk \left \{ \gamma^S  \begin{Vmatrix} \, R_k \mbf{x} - \tilde{\mbf{u}}_k  \end{Vmatrix}_{2}^{2} + \gamma^{LR} \begin{Vmatrix} \, R_k \mbf{x} -  \mbf{d}_k \end{Vmatrix}_{2}^{2} \right \} .
\end{align}
The normal equation of (\ref{eq:MRIsimpleForm}) for STROLLR-MRI in k-space is simplified to the following
\begin{align} \label{normal:MRI}
\nonumber  \begin{bmatrix} \Fb \Fbu^{H} \Fbu \Fb^{H} +  (\gamma^S + \gamma^{LR}) \Fb \sumi R_i^{*} R_i \Fb^{H} \end{bmatrix} \; &  \Fb \hat{\mbf{x}}  \\
= \Fb \Fbu^{H} \mbf{y} + \Fb \sumi R_i^{*} (\gamma^S \tilde{\mbf{u}}_i + \gamma^{LR} \vb_i) & \, .
\end{align}
On the left hand side of (\ref{normal:MRI}), matrix $\Fb \Fbu^H \Fbu \Fb^H = \diag{\boldsymbol{g}}$ is diagonal with binary entries equal to one at the sampled locations in k space, 
and $\Fb \sumi R_i^{*} R_i \Fb^{H} = n \Fb\, \Ib_p \Fb^{H} = n\, \Ib_p$ is a scaled identity. 
Therefore, a simple solution to (\ref{normal:MRI}) is
\begin{align} \label{sol:MRI}
\hat{\mbf{x}} = \Fb^{H} \Bb^{-1} \, \zb \, ,
\end{align}
where $\Bb \triangleq \diag{\boldsymbol{g}} + n (\gamma^S + \gamma^{LR}) \mbf{I}_p$ is a diagonal matrix whose inversion is cheap, and $\zb \triangleq \boldsymbol{G} \mbf{y} + \Fb \sumi R_i^H (\gamma^S \tilde{\mbf{u}}_i + \gamma^{LR} \vb_i)$, where $\boldsymbol{G} = \Fb \Fb_g^H  \in \mbb{C}^{p \times q}$ is a binary ``upsampling" matrix, which places the entries of $\mbf{y}$ in their corresponding $k$-space locations indicated by $\mbf{g}$ into a length-$p$ vector.



\section{Experiments} \label{sec5}

\begin{table*}[t]
\centering
\fontsize{8}{15pt}\selectfont
\begin{tabular}{|c|c|c|c|c|c|c|c|c|c|c|c|c|}
\hline
 & \multicolumn{5}{c|}{\textbf{Kodak} Dataset ($24$ images)}  & \multirow{2}{*}{$\Delta$PSNR}
 & \multicolumn{5}{c|}{\textbf{USC-SIPI Misc} Dataset ($44$ images)}  & \multirow{2}{*}{$\Delta$PSNR}
\\
\cline{1} \cline{2-6} \cline{8-12} 
$\sigma$ &          5 &         10 &         15 &         20 &         50 &  
 &          5 &         10 &         15 &         20 &         50 &  
\\
\hline

SF & 37.60 & 33.51 & 31.40 & 29.79 & 23.84 & -1.53
 & 36.93 & 33.27 & 31.40 & 30.01 & 24.08 & -2.66
\\
\hline

NLM & 36.85 & 32.91 & 30.93 & 29.62 & 25.55 & -1.59
 & 37.14 & 33.38 & 31.52 & 30.29 & 26.12 & -2.10
\\
\hline

GHP & 37.90 & 34.16 & 31.93 & 30.86 & 26.20 & -0.55
 & 36.32 & 33.36 & 31.55 & 30.59 & 26.42 & -2.14
\\
\hline

ODCT & 37.55 & 33.51 & 31.32 & 29.84 & 25.61 & -1.20
 & 38.12 & 34.27 & 32.16 & 30.71 & 26.35 & -1.47
\\
\hline

KSVD & 37.60 & 33.70 & 31.60 & 30.18 & 25.93 & -0.96
 & 38.33 & 34.67 & 32.69 & 31.35 & 26.95 & -1.00
\\
\hline

EPLL & 38.15 & 34.29 & 32.22 & 30.82 & 26.74 & -0.32
 & 38.41 & 34.67 & 32.67 & 31.32 & 27.13 & -0.95
\\
\hline

OCTOBOS & 38.27 & 34.24 & 32.16 & 30.70 & 26.48 & -0.39
 & 38.91 & 35.06 & 33.12 & 31.70 & 27.35 & -0.56
\\
\hline

PGPD & 38.21 & 34.37 & 32.32 & 30.92 & 27.08 & -0.18
 & 38.28 & 34.94 & 33.09 & 31.86 & 27.88 & -0.58
\\
\hline

BM3D & 38.30 & 34.39 & 32.30 & 30.92 & 26.98 & -0.18
 & 39.04 & 35.31 & 33.36 & 32.05 & 27.85 & -0.27
\\
\hline

NCSR & 38.35 & 34.48 & 32.36 & 30.92 & 26.84 & -0.17
 & 39.13 & 35.38 & 33.39 & 32.03 & 27.90 & -0.23
\\
\hline

SAIST & 38.39 & 34.51 & 32.39 & 30.98 & 26.95 & -0.12
 & 39.13 & 35.39 & 33.39 & 32.06 & 27.87 & -0.22
\\
\hline

\textbf{STROLLR} & $\mbf{38.46}$ & $\mbf{34.61}$ & $\mbf{32.50}$ & $\mbf{31.06}$ & $\mbf{27.18}$ & 0.00
 & $\mbf{39.26}$ & $\mbf{35.60}$ & $\mbf{33.63}$ & $\mbf{32.29}$ & $\mbf{28.18}$ & 0.00
\\
\hline
\end{tabular}
\vspace{0.1in}
\caption{Comparison of gray-scale image denoising PSNR values (in dB), averaged over the \textbf{Kodak} and \textbf{USC-SIPI Misc} datasets, using the proposed STROLLR image denoising method, versus other competing algorithms. For each dataset and noise level, the best denoising PSNR is marked in bold. For each method, $\Delta$ PSNR denotes the PSNR loss relative to the proposed STROLLR algorithm (highlighted in bold) averaged over the five different noise levels..}
\label{tab:denoisingDataset}
\vspace{-0.15in}
\end{table*}

\begin{figure}[tb]
\begin{center}
\begin{tabular}{ccc}
\hspace{-.12in}
\includegraphics[width=0.33\columnwidth]{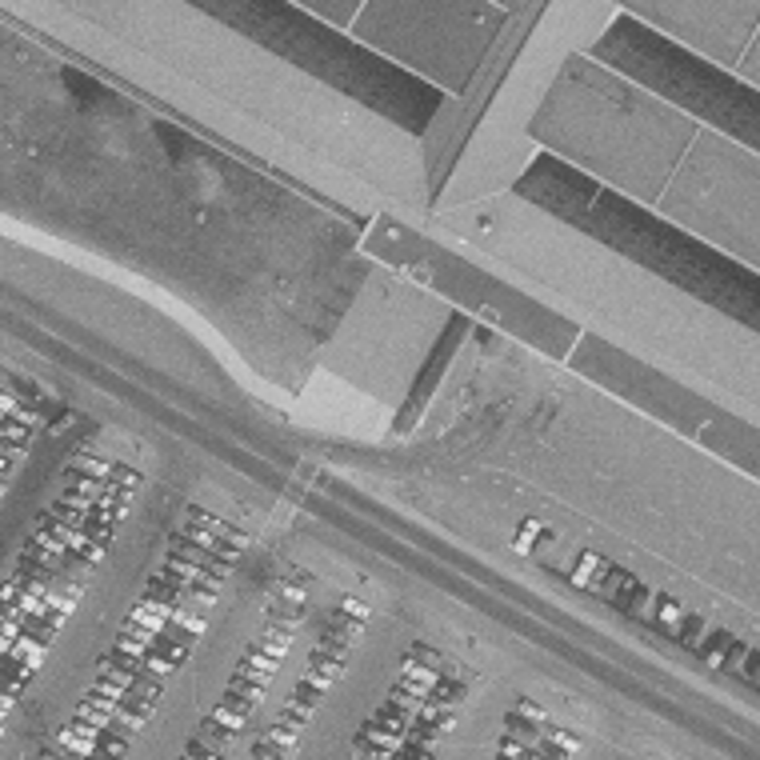} &
\hspace{-.2in}
\includegraphics[width=0.33\columnwidth]{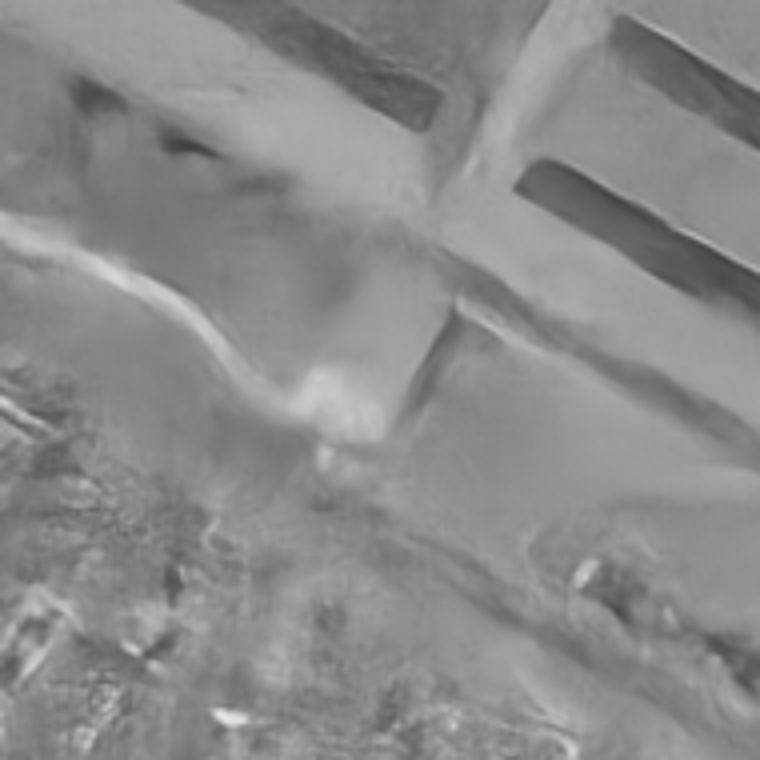} &
\hspace{-.2in}
\includegraphics[width=0.33\columnwidth]{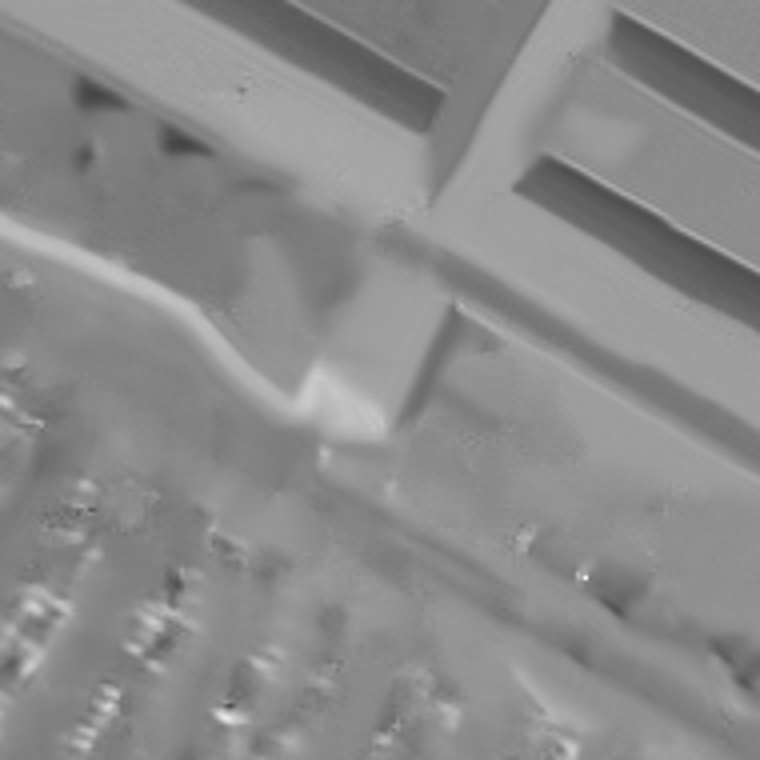} \\
\hspace{-.12in}
{\scriptsize (a) Ground Truth} &
\hspace{-.2in}
{\scriptsize (b) DnCNN ($25.87$ dB)} & 
\hspace{-.2in}
{\scriptsize (d) STROLLR ($\mbf{26.05}$ dB)} \\
\end{tabular}
\caption{Denoising result of the image \textit{Pentagon}: the zoom-in regions of (a) the ground truth, (b) the denoised image by DnCNN (PSNR = $25.87$ dB), and (c) the denoised image by STROLLR (PSNR = $\mbf{26.05}$ dB).}
\label{fig:denoisingDnCNN}
\end{center}
\vspace{-0.2in}
\end{figure}

\begin{figure*}
\begin{center}
\begin{tabular}{cccc}
\includegraphics[width=1.6in]{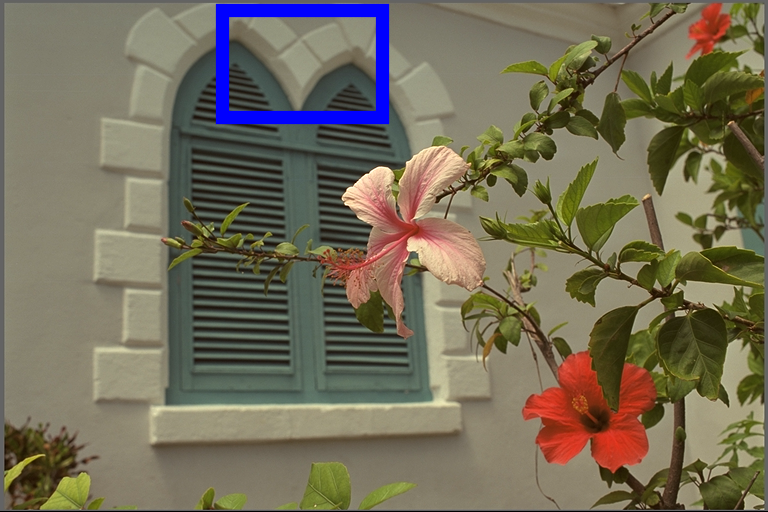} &
\includegraphics[width=1.6in]{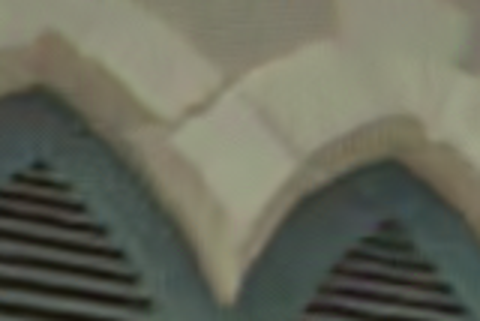} &
\includegraphics[width=1.6in]{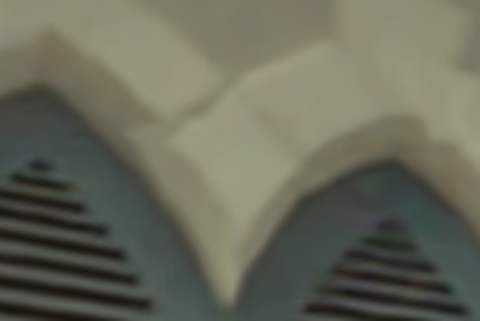} & 
\includegraphics[width=1.6in]{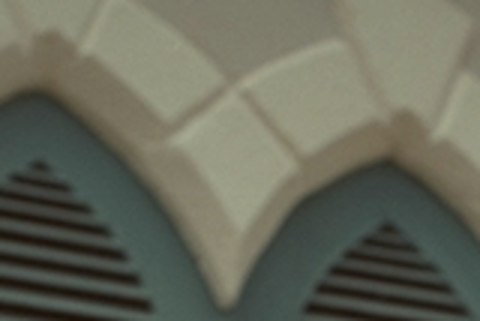} \\
\includegraphics[width=1.6in]{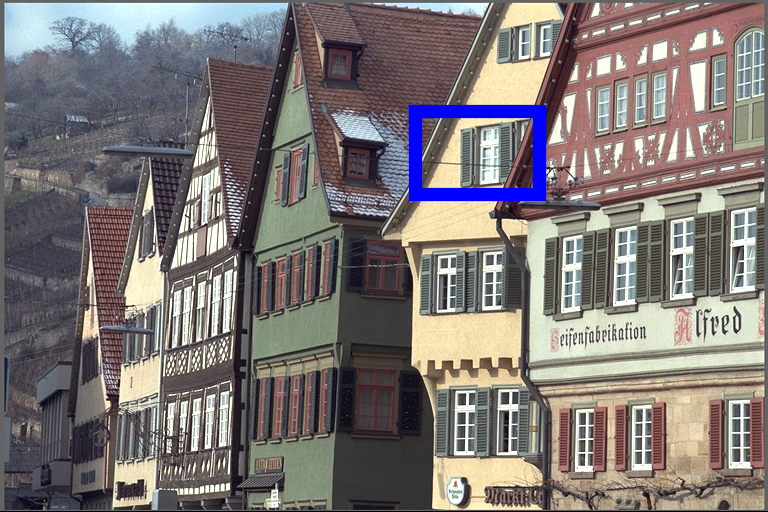} &
\includegraphics[width=1.6in]{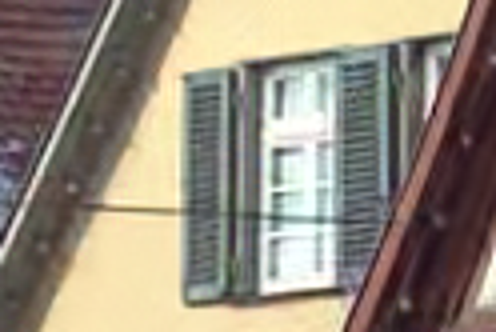} &
\includegraphics[width=1.6in]{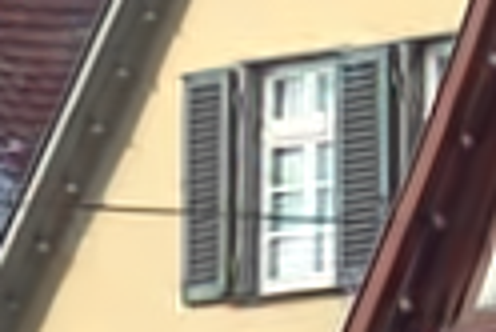} & 
\includegraphics[width=1.6in]{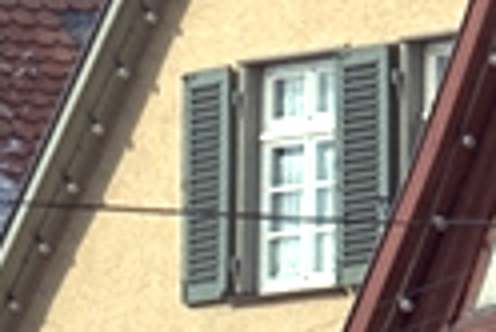}
\\
{\footnotesize (a) Zoom-in Regions} & {\footnotesize (b) C-BM3D: $31.64$ / $27.82$ dB} & {\footnotesize (c) C-STROLLR: $\mbf{32.08}$ / $\mbf{28.45}$ dB} & {\footnotesize (d) Ground Truth} 
\label{fig:denoisingColor}
\end{tabular}
\end{center}
\vspace{-0.1in}
\caption{Denoising results of (a) the example color images \textit{Kodim07} (top) and \textit{Kodim08} (bottom) at $\sigma = 35$, with the blue rectangles highlighting the zoom-in regions of (b) the images denoised by C-BM3D (PSNR = $31.64$ / $27.82$ dB), and (c) the images denoised by C-STROLLR (PSNR = $\mbf{32.08}$ / $\mbf{28.45}$ dB), and (d) the ground truth.}
\vspace{-0.1in}
\end{figure*}

In this section, we demonstrate the promise of the STROLLR based image restoration framework by testing our gray-scale / color image denoising, image inpainting, and CS MRI algorithms \footnote{The implementations of our proposed algorithms will be made publicly available upon acceptance of the paper.} on publicly available images or datasets \cite{kodak,sipi,Zhang2014,ravishankar2015efficient,wen2017frist,quData,sun2016deep}.
To evaluate the performance of the image recovery algorithms, we measure the peak signal-to-noise ratio (PSNR) in decibel (dB), which is computed between the ground true image and the recovered image.

All of the proposed STROLLR-based image recovery algorithms are unsupervised algorithms. 
\chg{There are several hyperparameters used in the algorithm, among which we set the spatial search window $\sqrt{Q} \times \sqrt{Q} = 30 \times 30$.
We initialize the unitary sparsifying transform $\Wb_0$ to be the 3D DCT (of size $\sqrt{n} \times  \sqrt{n}  \times l $).
These settings are fixed in all experiments.
}

\subsection{Image Denoising} \label{sec52}

We present image denoising results using our proposed algorithms in Sec.\ref{sec41}. 
For gray-scale image denoising experiment, we first convert the images in the \textit{Kodak} \cite{kodak} and \textit{SIPI Misc} \cite{sipi} datasets to gray-scale, and simulate i.i.d. Gaussian noise at 5 different noise levels ($\sigma = 5, 10, 15, 20$ and $50$). 
For the color image denoising experiment, we use all $24$ color images from the \textit{Kodak} dataset, and simulate equal-intensity i.i.d. Gaussian noise in each of the RGB channels at 4 different noise levels ($\sigma = 15, 25, 35$ and $50$).

\subsubsection{Implementation Details and Parameters}
\chg{We set the regularizer weights $\gamma^S = \gamma^{LR} = 1$, and the fidelity weight $\gamma^F = 0.1 / \sigma^2$, where $\sigma$ is the noise standard deviation of the noisy image $\mbf{y}$.
We directly use the noisy image as the initial estimate $\hat{\mbf{x}}_{0}$.
Only for denoising, 
we modify the image update step (\ref{eq:ImageUpdateDenoising}), at iteration $t = 1,2,...T-1$ (except for the last iteration), as follows
\begin{align} \label{eq:ImageUpdateDenoising2}
\hat{\mbf{x}}_t = \delta \Bb^{-1} \, \zb_t \, + (1-\delta) \mbf{y} \, ,
\end{align}
where we set $\delta = 0.1$. 
The modified $\hat{\mbf{x}}_t$ is the convex combination of the denoised estimate and the original noisy image.
This method, widely known as iterative regularization for non-convex inverse problems \cite{kaltenbacher2008iterative,gu2017weighted}, 
has been applied in various popular image restoration algorithms \cite{xu2015patch,xu2017multi,gu2017weighted,romano2015boosting}.
Let $\tilde{\mbf{x}}_t \triangleq \hat{\mbf{x}}_t - \mbf{x}$ denote the noise remaining in $\hat{x}_t$.
We re-estimate the variance of $\tilde{\mbf{x}}_t$ using $\sigma_t^2 = \psi (\sigma^2 - (1/N) \| \mbf{y} - \hat{\mbf{x}}_t \|^2)$~\cite{gu2017weighted,romano2015boosting}.
Here $(1/N) \| \hat{\mbf{x}}_t - \mbf{y} \|^2$ is an estimate of the variance of the noise removed throughout the $t$ iterations.
Assuming the removed noise to be white and uncorrelated with $\tilde{\mbf{x}}_t$, the estimated variance of $\tilde{\mbf{x}}_t$ is $\sigma^2 - (1/N) \| \hat{\mbf{x}}_t - \mbf{y} \|^2$.
However, because in practice the removed and the remaining noises are positively correlated, $\sigma_t^2$ tends to be over-estimated using the ideal formula.
To better approximate the actual $\sigma_t^2$, 
we compensate by the factor $\psi = 0.36$~\cite{xu2015patch,xu2017multi,gu2017weighted,romano2015boosting}.
At the $t$th iteration, we set the penalty parameters $\lambda = 1.2 \sigma_{t-1}$ and $\theta = 0.8 \sigma_{t-1} (\sqrt{n} + \sqrt{M})$~\cite{wen2017joint}, using the re-estimated $\sigma_{t-1}$ (or the noise level of $\mbf{y}$, $\sigma_0 = \sigma$ at the first iteration).
The remaining hyper parameters are the patch size $n$, data matrix sizes $M$ and $l$, and the number of iterations $T$.
We set them to be $\begin{Bmatrix} n,M,l,T \end{Bmatrix} = \begin{Bmatrix} 6^2, 70, 8, 8 \end{Bmatrix}$ and $\begin{Bmatrix} 7^2, 80, 7, 10 \end{Bmatrix}$, for low-noise case (i.e., $0 \leq \sigma \leq 30$), and high-noise case (i.e. $\sigma > 30$), respectively.
}

\subsubsection{Gray-Scale Image Denoising}
We compare our proposed STROLLR based image denoising algorithm to various well-known alternatives, including denoising algorithms using overcomplete DCT (ODCT) dictionary, KSVD \cite{elad2}, GHP \cite{zuo2013texture}, Shrinkage Fields (SF) \cite{schmidt2014shrinkage}, EPLL \cite{zoran2011learning}, NLM \cite{Buades2005}, OCTOBOS \cite{octobos}, BM3D \cite{Dabov2007}, NCSR \cite{dong2013nonlocally}, PGPD \cite{xu2015patch}, and SAIST \cite{Dong2013}.
We use their publicly available codes for implementation. 
Among these methods, ODCT, KSVD and OCTOBOS exploit sparsity of image patches.
EPLL and GHP make use of image pixel statistics.
SF uses an unrolled neural network based on an analysis sparse model.
NLM, BM3D, SAIT, NCSR, and PGPD are all non-local methods that use collaborative filtering, low-rank, or sparse approximation.
Additionally, to better understand the benefit of each of the regularizers used in STROLLR model, we evaluate the denoising results using only the transform learning (TL), and the low-rank approximation (LR). 

Table \ref{tab:denoisingDataset} lists the denoised PSNRs obtained using the aforementioned methods, with the best result for each noise level and testing dataset (i.e., each column) marked in bold. 
The proposed STROLLR image denoising method provides average PSNR improvements of $0.2$dB, $0.2$dB, $0.3$dB, $0.4$dB, $0.5$dB, $0.7$dB, $1.0$dB, $1.4$dB, $1.6$dB, $2.0$dB, and $2.2$dB, respectively, over the SAIST, NCSR, BM3D, PGPD, OCTOBOS, EPLL, KSVD, ODCT, GHP, NLM and SF denoising methods. 
By imposing both sparsity and non-local (i.e., group low-rankness) regularizers, for all noise $\sigma$'s and testing datasets, STROLLR performs consistently the best. 
Thus our proposed method demonstrates robust and promising performance in image denoising compared to popular competing methods.

To further analyze the effectiveness of imposing both the sparsity and the low-rankness regularizers, as well as applying the iterative regularization method, we conduct an ``ablation" study by disabling in turn each of these three components in the STROLLR image denoising algorithm.
We run STROLLR denoising with single pass, i.e., without iterative regularization, which is denoted as STROLLR-S.
On top of STROLLR-S, we further disable the low-rank regularizer $\Rf_{LR}$, and the sparsity regularizer $\Rf_{S}$, which are referred as STROLLR-S w/o LR, and STROLLR-S w/o TL, respectively.
Table \ref{tab:ablation} lists the denoised PSNRs obtained using STROLLR, and its variants. 
It is clear that STROLLR denoising outperforms its variants, and all of the three components contribute significantly to the success of the proposed STROLLR algorithm.

\begin{table}
\centering
\fontsize{8}{15pt}\selectfont
\begin{tabular}{|c|c|c|c|c|c|}
\hline
 & \multicolumn{4}{c|}{\textbf{Kodak} Dataset ($24$ images)}  & $\Delta$
\\
\cline{1-5} 
$\sigma$ & 5 &  10  &   20  &  50 &  PSNR
\\
\hline
{\scriptsize STROLLR-S w/o TL} & 38.15 & 34.09 & 30.33 & 25.82 & -0.73
\\
\hline
{\scriptsize STROLLR-S w/o LR} & 38.09 & 34.02 & 30.36 & 25.96 & -0.72 
\\
\hline
STROLLR-S & 38.31 & 34.43 & 30.96 & 26.87 & -0.19
\\
\hline
STROLLR & $\mbf{38.46}$ & $\mbf{34.61}$ & $\mbf{31.06}$ & $\mbf{27.18}$ & 0
\\
\hline
\end{tabular}
\vspace{0.1in}
\caption{PSNRs of gray-scale image denoising, using STROLLR and its variants, averaged over the \textbf{Kodak} image dataset. For each noise level, the best denoising PSNR is marked in bold. For each variant, $\Delta$PSNR denotes the PSNR loss relative to the full STROLLR denoiser, averaged over the four noise levels.}
\label{tab:ablation}
\vspace{-0.15in}
\end{table}

\begin{table}
\centering
\fontsize{8}{15pt}\selectfont
\begin{tabular}{|c|c|c|c|c|c|}
\hline
 & \multicolumn{4}{c|}{\textbf{Kodak} Dataset ($24$ color images)}  & \multirow{2}{*}{$\Delta$PSNR}
\\
\cline{1-5} 
$\sigma$ & 15 &   25  &     35 &    50 &  
\\
\hline
WNNM & 32.49 & 29.68 & 28.49 & 26.93 & $-1.98$
\\
\hline
TNRD & N/A& 30.08 & N/A & 27.17 & $-1.76$
\\
\hline
MC-WNNM & 33.94 & 31.35 & 29.70 & 28.02 & $-0.63$
\\
\hline
C-BM3D & 34.41 & 31.81 & 30.04 & 28.62 & $-0.16$
\\
\hline
C-STROLLR & $\mbf{34.57}$ & $\mbf{31.94}$ & $\mbf{30.25}$ & $\mbf{28.78}$ & $0$
\\
\hline
\end{tabular}
\vspace{0.1in}
\caption{PSNR values of color image denoising, averaged over the \textbf{Kodak} color image dataset, using the TNRD \cite{chen2017trainable}, MC-WNNM \cite{xu2017multi}, C-BM3D \cite{dabov2007color}, and the proposed C-STROLLR image denoiser. For each noise level, the best denoising PSNR is marked in bold. For each method, $\Delta$ PSNR denotes the PSNR loss relative to the proposed STROLLR algorithm (highlighted in bold), averaged over the four noise levels.}
\label{tab:colorDenoisingDataset}
\vspace{-0.25in}
\end{table}

Besides the conventional approaches, popular deep learning techniques \cite{ye2018deep,bae2017beyond,chen2017trainable,schmidt2014shrinkage,burger2012image} have been shown useful in image denoising. 
The recently proposed DnCNN \cite{zhang2017beyond} demonstrated superior image denoising results over the standard natural image datasets. 
The deep learning based methods typically require a large training corpus containing images that have similar distribution to the image to be recovered. 
However, such a training corpus may not always be easy to obtain in applications such as remote sensing, biomedical imaging, etc. 
Fig. \ref{fig:denoisingDnCNN} shows an example denoising result of the image \textit{Pentagon} from the SIPI aerial dataset \cite{sipi}. 
Here we applied the DnCNN algorithm using the publicly available implementation and the trained models (using 400 images from BSDS500 dataset \cite{roth2009fields}) from the authors' project website.
Comparing to the denoised result using the proposed STROLLR-based algorithm, DnCNN generates various artifacts and distortions in the denoised image.

\subsubsection{Color Image Denoising}
The C-STROLLR algorithm extends STROLLR to color image denoising as described before. 
We compare to popular denoising methods including WNNM \cite{gu2017weighted}, TNRD \cite{chen2017trainable}, MC-WNNM \cite{xu2017multi}, and C-BM3D \cite{dabov2007color}. 
Among the selected competing methods, TNRD is a deep learning based method, while WNNM, MC-WNNM, and C-BM3D are all internal methods using non-local image structures.
WNNM is based on low-rank approximation, which is a gray-scale image denoising algorithm. Thus, we apply WNNM to each of the RGB channels of color images.
MC-WNNM is the color image denoising extension of WNNM algorithm, which further exploits the cross-channel correlation.

Table \ref{tab:colorDenoisingDataset} lists the average denoising PSNRs over the $24$ color images from \textit{Kodak} database, with the best result for each noise level marked in bold. 
It is clear that the proposed C-STROLLR performs consistently the best for all noise levels. 
The MC-WNNM algorithm generates significantly better results comparing to those using the channel-wise WNNM denoising, which demonstrates the importance of exploiting image color correlation in restoration.
Fig. \ref{fig:denoisingColor} compares the denoising results of example images \textit{Kodim07} and \textit{Kodim08} at $\sigma = 35$, using the best competitor C-BM3D, and the proposed C-STROLLR algorithms.
The denoised image by C-STROLLR preserves clearer details thanks to the sparsity regularization, while C-BM3D generates undesired artifacts, e.g., the zoomed-in region in the blue boxes.
We observe similar, or more severe artifacts in the denoising results using other competing methods.

\begin{table}
\centering
\fontsize{7}{13pt}\selectfont
\begin{tabular}{|c|c|c|c|c|c|c|}
\hline
Image & \multicolumn{3}{c|}{\textit{Barbara}}  & \multicolumn{3}{c|}{\textit{House}}
\\
\hline
Available & \multirow{2}{*}{$10\%$} & \multirow{2}{*}{$20\%$} & \multirow{2}{*}{$50\%$} & \multirow{2}{*}{$10\%$} & \multirow{2}{*}{$20\%$} & \multirow{2}{*}{$50\%$}
\\
pixels & & & & & & 
\\
\hline
Bicubic & 22.65 & 23.65 & 25.94 & 29.82 & 31.62 & 33.18
\\
\hline
SKR & 21.92 & 22.45 & 22.81 & 30.18 & 31.05 & 31.94
\\
\hline
Smooth & 28.32 & 30.90 & 35.94 & 33.67 & 36.62 & 39.98
\\
\hline
GSR & 31.32 & 34.42 & 39.12 & 35.61 & 37.65 & 41.61
\\
\hline
STROLLR & $\mbf{31.51}$ & $\mbf{34.56}$ & $\mbf{39.33}$ & $\mbf{35.72}$ & $\mbf{37.75}$ & $\mbf{41.70}$
\\
\hline
\end{tabular}
\vspace{0.1in}
\caption{PSNR values of image inpainting, using bicubic interpolation, SKR, GSR, and the proposed STROLLR image inpainting method. For each image and available pixel percentage, the best inpainting PSNR is marked in bold.}
\label{tab:imageInpainting}
\vspace{-0.25in}
\end{table}

\subsection{Image Inpainting} \label{sec53}

\begin{table*}[t]
\begin{center}
\fontsize{9}{14pt}\selectfont
\begin{tabular}{|c|c|c|c|c|c|c|c|c|c|c|}
\hline
\multirow{2}{*}{Image}  &  Sampling & Under- & Zero- & Sparse & DL- & \multirow{2}{*}{PBDWS} & \multirow{2}{*}{PANO} & TL- & FRIST- & STROLLR- \\
 & Scheme & sampl. & filling & MRI & MRI & & & MRI & MRI  & MRI\\
\hline
a & 2D Random & $5 \times $ & $26.9$ & $27.9$ & $30.5$ & $30.3$ & $30.4$ & $30.6$ & $30.7$ & \textbf{32.4} \\  
\hline
b & Cartesian & $2.5 \times $ & $28.1$ & $31.7$ & $37.5$ & $42.5$ & $40.0$ & $40.7$ & $40.9$ & \textbf{44.0} \\  
\hline
c & Cartesian & $2.5 \times $ & $24.9$ & $29.9$ & $36.6$ & $35.8$ & $34.8$ & $36.3$ & $36.7$ & \textbf{40.9} \\  
\hline
\multirow{4}{*}{d} & Cartesian & $4 \times $ & $28.9$ & $29.7$ & $32.7$ & $31.7$ & $32.7$ & $32.8$ & $33.0$ & \textbf{35.2} \\  
\cline{2-11}
 & Cartesian & $7 \times $ & $27.9$ & $28.6$ & $30.9$ & $31.1$ & $31.1$ & $31.2$ & $31.4$ & \textbf{32.8} \\  
\cline{2-11}
 & 2D Random & $4 \times $ & $25.2$ & $26.1$ & $33.0$ & $31.7$ & $32.8$ & $33.1$ & $33.1$ & \textbf{35.6} \\  
\cline{2-11}
 & 2D Random & $7 \times $ & $25.3$ & $26.4$ & $31.7$ & $31.1$ & $30.9$ & $31.9$ & $32.1$ & \textbf{33.3} \\  
\hline
\end{tabular}
\caption{PSNRs, corresponding to the Zero-filling, Sparse MRI \cite{lustig2007sparse}, DL-MRI \cite{ravishankar2011mr}, PBDWS \cite{ning2013magnetic}, PANO \cite{qu2014magnetic}, TL-MRI \cite{ravishankar2015efficient}, FRIST-MRI \cite{wen2017frist}, and the proposed STROLLR-MRI reconstructions for various images, sampling schemes, and undersampling factors. The best PSNR for each case is marked in bold.}
\label{table:mri}
\end{center}
\vspace{-0.3in}
\end{table*}

We present the image inpainting results using our STROLLR based algorithm described in Sec.\ref{sec42}. 
The testing gray-scale images \textit{Barbara} and \textit{House} which were used in the GSR inpainting method \cite{Zhang2014} are selected to evaluate our proposed STROLLR method, as well as competing methods including Bicubic interpolation, SKR \cite{takeda2007kernel}, Smooth \cite{ram2013image}, and GSR \cite{Zhang2014}.
\chg{
We work with $6 \times 6$ patches, and set $M = 80$, $l = 8$, the number of iterations $T = 150$, and $\gamma^S = \gamma^{LR} = 1$.
We set the rank penalty weight to be  $\theta = \lambda  (\sqrt{n} + \sqrt{M})$.
We randomly remove image pixels, and keep only $20\%$, $30\%$ and $50\%$ pixels of the entire image, and set the sparsity penalty weight $\lambda = 20, 12$ and $5$ respectively.
}

Table \ref{tab:imageInpainting} lists the inpainting PSNRs, over all available pixel percentages and testing images, obtained using the aforementioned methods. The best result for each testing case is marked in bold. 
The proposed STROLLR inpainting algorithm produces better results than the popular competitors.

\subsection{MRI reconstruction} 

\begin{figure}
\begin{center}
\begin{tabular}{ccc}
\hspace{-0.1in}
\includegraphics[width=1.1in]{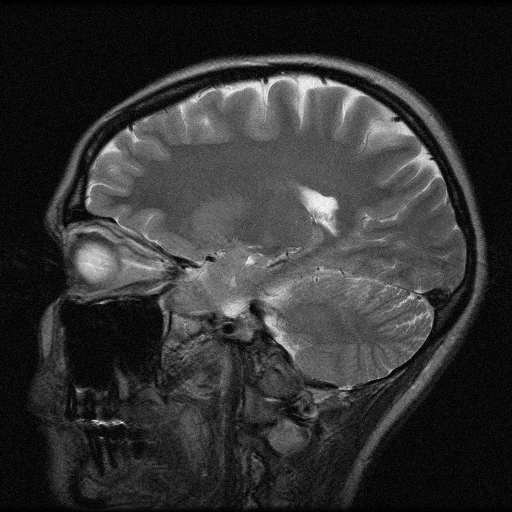} &
\hspace{-0.15in}
\includegraphics[width=1.1in]{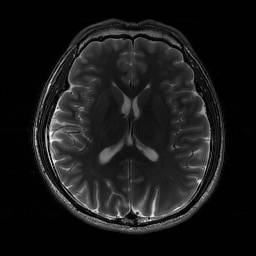} & 
\hspace{-0.15in}
\includegraphics[width=1.1in]{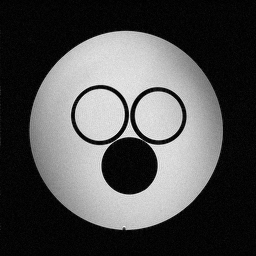}
\\
\hspace{-0.1in}
{\small (a)} & 
\hspace{-0.15in}
{\small (b)} & 
\hspace{-0.15in}
{\small (c)}
\end{tabular}
\end{center}
\vspace{-0.1in}
\caption{Tesing anatomical and physical phantom MR images: (a) is used in \cite{ravishankar2015efficient,wen2017frist}, and (b) and (c) are from a publicly available dataset \cite{quData}.}
\label{fig:MRIimages}
\vspace{-0.1in}
\end{figure}

\begin{figure}
\begin{center}
\begin{tabular}{ccc}
\hspace{-0.15in}
\includegraphics[width=1.1in]{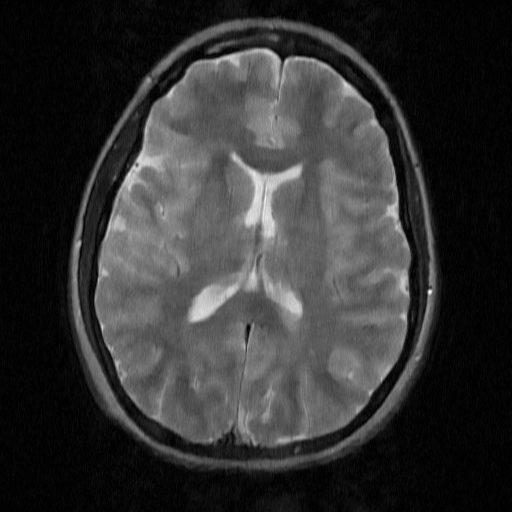} &
\hspace{-0.2in}
\includegraphics[width=1.1in]{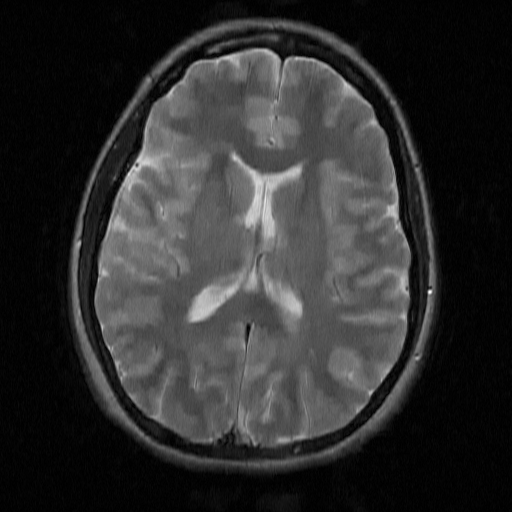} & 
\hspace{-0.2in}
\includegraphics[width=1.1in]{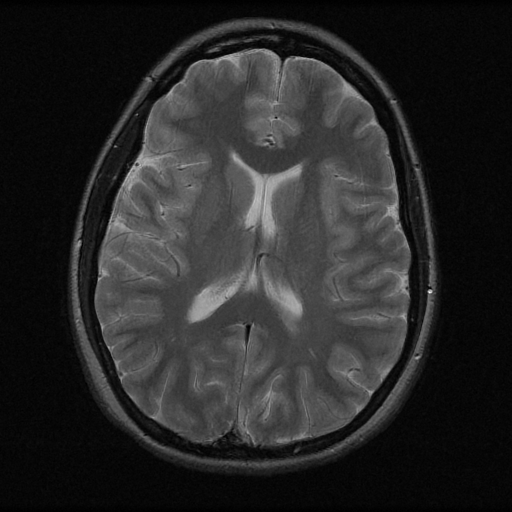} 
\\
\vspace{-0.1in}
\hspace{-0.15in}
\includegraphics[width=1.1in]{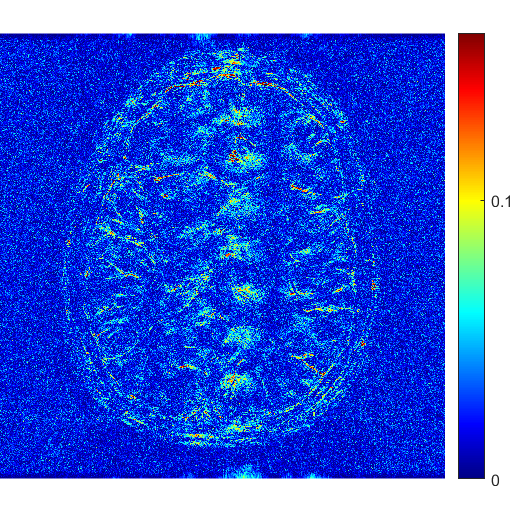} &
\hspace{-0.2in}
\includegraphics[width=1.1in]{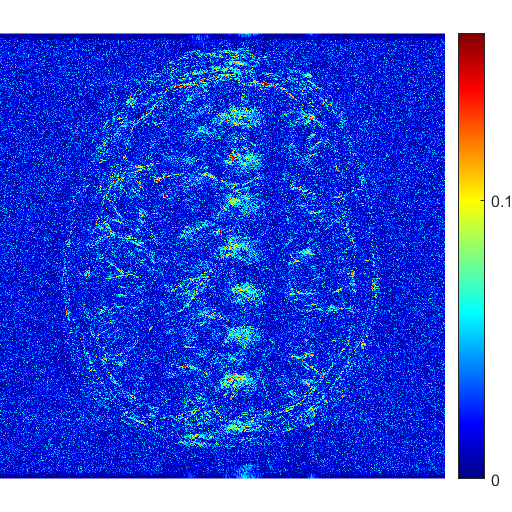} & 
\hspace{-0.2in}
\includegraphics[width=1.1in]{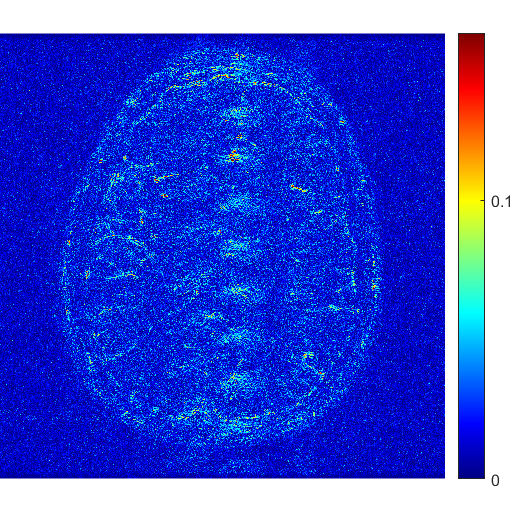} 
\\
\hspace{-0.15in}
{\scriptsize (a) DL-MRI, $31.7$dB} & 
\hspace{-0.2in}
{\scriptsize (b) FRIST-MRI, $32.1$dB} & 
\hspace{-0.2in}
{\scriptsize (c) STROLLR-MRI $\mbf{33.3}$dB} 
\end{tabular}
\end{center}
\vspace{-0.1in}
\caption{Reconstruction of MR image \textbf{d} using (a) DL-MRI, (b) FRIST-MRI, and (c) the proposed STROLLR-MRI. Top row: reconstructions; bottom row: magnitude of the reconstruction error.}
\label{fig:MRIresults}
\vspace{-0.15in}
\end{figure}

\new{We present the MRI reconstruction results using the proposed STROLLR-MRI algorithm. 
The $4$ testing MR images (3 anatomical and one physical phantom), i.e., the images \textbf{a} - \textbf{c} used in our experiments shown in Fig. \ref{fig:MRIimages}(a)-(c) and the image \textbf{d} shown in Fig. \ref{fig:MRIresults}, are all publicly available \cite{ravishankar2015efficient,wen2017frist,quData}.
We simulated complex MR data obtained by taking the discrete Fourier transform (DFT) of the magnitude of the complex images \footnote{The STROLLR-MRI, as well as the competing methods except for ADMM-Net~\cite{sun2016deep} can handle complex MR data.}, with various undersampling masks in k-space, using either Cartesian or 2D random sampling patterns \cite{trzasko2009highly,ravishankar2015efficient}, at undersampling ratios ranging from $2.5\times$ to $7\times$.}
The proposed STROLL-MRI scheme is applied to reconstruct the MR images.
We set the weights of the STROLLR regularizers to $\gamma^S = \gamma^{LR} = 10^{-6}$.
The sparsity and low-rankness penalty coefficients coefficients $\theta = 2\lambda = \theta_0$, where $\theta_0$ depends on undersampling ratio of the k-space measurement, as well as the image distribution. 
\new{For the three anatomical images \textbf{a}, \textbf{b} and \textbf{d}, we set $\theta_0 = 0.02$ when the undersampling ratio is smaller or equal to $5\times$, and $\theta_0 = 0.05$ when undersampling ratio is higher than $5\times$.
For the physical phantom image \textbf{c}, which has large piece-wise smooth regions, we set $\theta_0 = 0.05$ for reconstructing $2.5\times$ undersampled k-space measurement.}
We run STROLLR-MRI for $100$ iterations, and have observed the empirical convergence of the objective functions.

We first compare our STROLLR-MRI reconstruction results to those obtained using popular \textit{internal} methods, including naive Zero-filling, Sparse MRI \cite{lustig2007sparse}, PBDWS \cite{ning2013magnetic}, PANO \cite{qu2014magnetic}, DL-MRI \cite{ravishankar2011mr}, TL-MRI \cite{ravishankar2015efficient}, and FRIST-MRI \cite{wen2017frist}. 
To evaluate the performance of the MRI reconstruction schemes, we measure the reconstruction PSNRs (computed for image magnitudes) for various approaches, which are listed in Table \ref{table:mri}.
The proposed STROLLR-MRI algorithm provides PSNR improvements (averaged over all 7 cases) of $2.3$dB, $2.5$dB, $3.1$dB, $2.9$dB, $3.0$dB, $7.7$dB, over the FRIST-MRI, TL-MRI, PANO, PBDWS, DL-MRI, and Sparse MRI, respectively. 
Even when compared to the recently proposed TL-MRI and FRIST-MRI, our STROLLR-MRI provides much better reconstruction results. 
The quality improvement obtained by STROLLR-MRI demonstrates the effectiveness of the learned STROLLR as the regularizer, as it utilizes both sparsity and non-local similarity. 
Figure \ref{fig:MRIresults} compares the reconstructed MR images, and the magnitudes of their reconstruction error (i.e., difference between the magnitudes of the reconstructed images and the reference images) using \new{DL-MRI, FRIST-MRI, and STROLLR-MRI.}
The reconstruction result using STROLLR-MRI contains fewer artifacts and less noise, compared to competing methods.

\begin{figure}
\begin{center}
\begin{tabular}{ccc}
\hspace{-0.15in}
\includegraphics[width=1.1in]{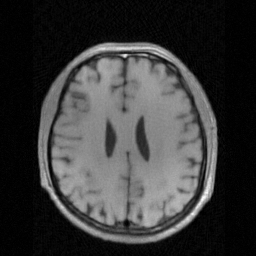} &
\hspace{-0.2in}
\includegraphics[width=1.1in]{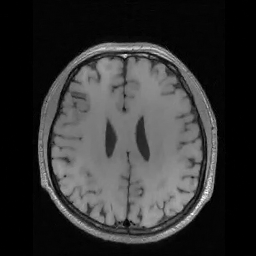} & 
\hspace{-0.2in}
\includegraphics[width=1.1in]{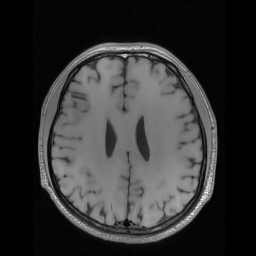} \\
\vspace{-0.1in}
\hspace{-0.15in}
\includegraphics[width=1.1in]{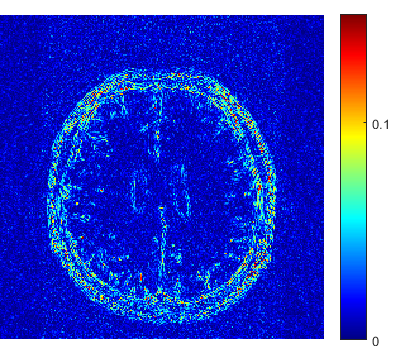} &
\hspace{-0.2in}
\includegraphics[width=1.1in]{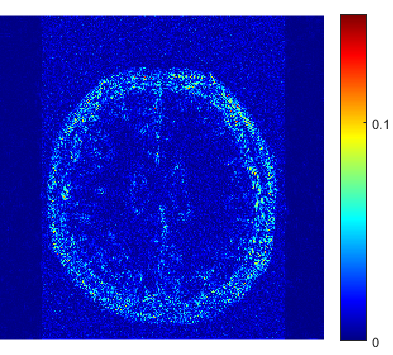} & 
\hspace{-0.2in}
\includegraphics[width=1.1in]{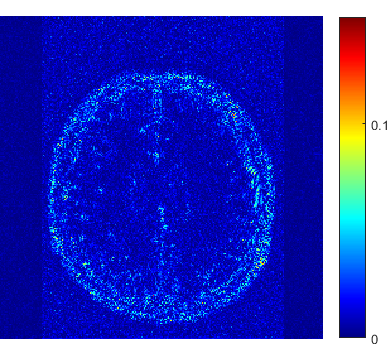}
\\
\hspace{-0.25in}
{\scriptsize (a) TL-MRI, $32.4$dB} & 
\hspace{-0.25in}
{\scriptsize (b) ADMM-Net, $35.9$dB} & 
\hspace{-0.23in}
{\scriptsize (c) STROLLR-MRI $\mbf{37.3}$dB} 
\end{tabular}
\end{center}
\vspace{-0.1in}
\caption{Example of CS MRI of the \textit{Brain Data $1$} using the $5\times$ undersampled pseudo radial mask in K-space. Reconstructions using (a) TL-MRI, (b) ADMM-Net, and (c) the proposed STROLLR-MRI (top row), and the magnitudes of the corresponding reconstruction error (bottom row).}
\label{fig:MRIcompareADMM}
\vspace{-0.15in}
\end{figure}

Finally, we compare the proposed STROLLR-MRI scheme to the recent ADMM-Net \cite{sun2016deep}, which is an \textit{external} method using deep learning.
We use the publicly available implementation with the trained $15$-stage model by Yang et al \cite{sun2016deep}.
Because the released ADMM-Net model requires the MR image to have fixed size (i.e., $256 \times 256$) with a specific sampling mask (i.e., pseudo radial sampling pattern with $5\times$ undersampling ratio) \cite{sun2016deep}, the testing images \textbf{a}-\textbf{d} are not applicable for direct comparison.
Instead, we reconstruct the two example MR images that were used in the ADMM-Net demonstration \cite{sun2016deep}.
The proposed STROLLR-MRI algorithm provides PSNR improvements of $1.4$dB and $1.3$dB over ADMM-Net, for reconstructing the testing image \textit{Brain data $1$} and \textit{Brain data $2$} , respectively.
\new{Figure \ref{fig:MRIcompareADMM} compares the reconstructed MR images with the magnitudes of their corresponding reconstruction errors using TL-MRI \cite{ravishankar2015efficient}, ADMM-Net \cite{sun2016deep}, and STROLLR-MRI.}
It is clear that the reconstruction result using STROLLR-MRI provides less noise and artifacts comparing to ADMM-Net.
Comparing to deep learning algorithms, 
the proposed STROLLR-MRI does not require re-training the model using different sampling masks, over an image corpus with similar distribution, when reconstructing a class of MR images.



\section{Conclusion} \label{sec6}

We presented an effective image restoration framework using transform learning scheme with joint low-rank regularization.
The novel STROLLR model exploits both the image sparisity in the learned sparsifying transform, and the image nonlocal self-similarity using group low-rankness.
We proposed several efficient algorithms for image denoising, inpainting, and CS MRI applications.
We demonstrated the promising performance of the STROLLR image restoration methods for some publicly available datasets.
Our proposed algorithms outperformed all competing schemes, including the popular, and the state-of-the-art methods.
While this work demonstrated the promise of the proposed STROLLR model, we plan to systematically study the combination of various effective regularizers in inverse problems in future work.

\ifCLASSOPTIONcaptionsoff
  \newpage
\fi

\bibliographystyle{./IEEEtran}
\bibliography{./refs}

\end{document}